\documentclass[11pt]{article}

\usepackage[preprint]{acl}

\usepackage{times}
\usepackage{latexsym}

\usepackage[T1]{fontenc}

\usepackage[utf8]{inputenc}

\usepackage{microtype}

\usepackage{inconsolata}

\usepackage{graphicx}
\usepackage{enumitem}
\usepackage{amsmath}
\usepackage{paralist}
\usepackage{booktabs} 
\usepackage{multirow}
\usepackage{makecell}
\usepackage[table]{xcolor}

\usepackage{tcolorbox}
\usepackage{subcaption}
\usepackage{tabularx}
\usepackage{dashrule}
\usepackage{colortbl}
\usepackage{arydshln}

\newcommand{\nop}[1]{}

\title{Does Memory Need Graphs? A Unified Framework and Empirical Analysis for Long-Term Dialog Memory}

\author{
Sen Hu\textsuperscript{1,$\dagger$,*},
Yuxiang Wei\textsuperscript{2,$\dagger$},
Jiaxin Ran\textsuperscript{1,$\dagger$},\\
\textbf{Zhiyuan Yao}\textsuperscript{3},
\textbf{Xueran Han}\textsuperscript{4},
\textbf{Huacan Wang}\textsuperscript{5},
\textbf{Ronghao Chen}\textsuperscript{1},
\textbf{Lei Zou}\textsuperscript{1} \\[0.5ex]
\textsuperscript{1}Peking University \quad
\textsuperscript{2}Georgia Tech \quad
\textsuperscript{3}ZJU \quad
\textsuperscript{4}MBZUAI \quad
\textsuperscript{5}UCAS \\[1.5ex]
\textsuperscript{$\dagger$}Equal contribution \quad
\textsuperscript{*}Corresponding author: \texttt{husen@pku.edu.cn}
}

\begin{document}
\maketitle
\begin{abstract}

Graph structures are increasingly adopted in dialog memory systems, motivated by their success in retrieval-augmented generation and the associative nature of human memory. However, empirical findings on their effectiveness remain inconsistent, making it unclear which design choices truly matter. 
In this work, we present an experimental and system-oriented analysis of long-term dialog memory architectures. We formalize a unified framework that decomposes dialog memory systems into core components and supports both graph-based and non-graph approaches. Under this framework, we conduct controlled, stage-wise experiments on LongMemEval and HaluMem, comparing common design choices in memory representation, organization and maintenance, as well as indexing and retrieval. 
Our results show that underlying implementation details—often insufficiently specified in prior work—have a substantial impact on performance, and we identify stable, reliable strong baselines to support fair comparison and practical deployment.
Code are available at \url{https://github.com/AvatarMemory/UnifiedMem}

\end{abstract}

\section{Introduction}

With the widespread adoption of graph structures in retrieval-augmented generation (RAG) systems \citep{peng2024graph,zhang2025survey}, and the rapid development of research on long-term dialog memory, incorporating graph structures into dialog memory systems has gradually become a common design choice \citep{hu2025memory}.
From an intuitive perspective, human memory is not stored as isolated units but organized through associative relationships \citep{wu2025human}, which makes explicit relational modeling a natural conceptual fit for dialog memory.

Despite this intuition, existing literature reveals that empirical conclusions on dialog memory methods remain far from consistent.
Some studies \cite{chhikara2025mem0, zhang2025bridging} report performance gains from introducing relatively complex graph construction and graph-based retrieval mechanisms, while other works \cite{wu2025sgmem, fang2025lightmem} show that non-graph paradigms or lightweight structures achieve comparable or even better results on the same or similar benchmarks.
In the absence of a unified analytical perspective, such discrepancies are difficult to attribute to specific design choices, which complicates both the interpretation of prior results and the practical decision-making process for system builders.

We observe that comparing and attributing existing results remains difficult for several reasons.
Different works often adopt different datasets, backbone models, or evaluation settings, and underlying implementation details are sometimes insufficiently specified, limiting reproducibility.
More fundamentally, dialog memory research lacks a unified system-level framework that covers both graph-based and non-graph approaches.

As a result, method-centric studies often describe end-to-end systems around a small set of highlighted innovations, while other system components—such as memory representation, indexing, retrieval and ranking—are implemented with varying assumptions and levels of specificity across methods.
When these foundational settings are not aligned, performance differences become difficult to attribute to specific design choices, and reported results across studies can be difficult to interpret or reconcile~\footnote{More details can be found in Appendix \ref{appendix_relatedsystem}}.

Based on these observations, this paper is positioned as an \textbf{experimental and system-oriented analysis} rather than a proposal of a new method.
Our contributions are summarized as follows:
\begin{compactitem}
\item Inspired by LongMemEval, we formalize a unified framework for dialog memory that encompasses both graph-based and non-graph methods, providing a foundation for aligning and comparing different approaches.

\item Using the LongMemEval and HaluMem benchmarks, we conduct controlled, stage-wise experimental comparisons of several commonly adopted design choices in dialog memory systems, covering memory unit design, memory organization and maintenance mechanisms, as well as graph-based and non-graph indexing and retrieval strategies.

\item Through fine-grained empirical analysis, we identify and validate a set of stable and reliable strong baselines that hold across both graph-based and non-graph approaches. These baselines can serve as a common starting point for future dialog memory research and system development, lowering the barrier to fair comparison and practical deployment.
\end{compactitem}

\section{Related Work} \label{sec:related}

\subsection{Retrieval-Augmented Generation}

Retrieval-Augmented Generation (RAG) retrieves relevant information from a pre-constructed external knowledge base to enrich the prompt, thereby enabling large language models to produce more grounded and accurate responses~\cite{lewis2020retrieval}. 

To address these limitations, recent work has introduced graph-structured knowledge into the retrieval process. GraphRAG~\cite{edge2024local} leverages graph topology to capture rich entity relationships, enabling relationship-driven retrieval and multi-hop reasoning. Building on this line of work, HippoRAG2~\cite{gutierrez2025rag} adopts a hybrid hierarchical graph and performs structure-aware retrieval via graph-based propagation. LightRAG~\cite{guo2024lightrag} proposes a dual-level graph retrieval mechanism that enhances coverage and reasoning by jointly leveraging fine-grained local knowledge and higher-level global structure. Collectively, these methods point toward a more structured direction for future memory systems in intelligent agents, improving both knowledge organization and reasoning capability.

\subsection{Agent Memory Augmentation}

Memory plays a crucial role in agent systems, serving as a key component that enables agents to sustain long-term attention, dynamically acquire new knowledge from historical data, and act autonomously. In the context of long-term dialog agents, research on memory augmentation can be broadly divided into two categories: plain-index-based and graph-index-based methods. Wu et al. experimentally demonstrated that existing LLMs and commercial chat assistants exhibit significant performance degradation in handling long‑term dialog memory, leading them to propose the LME benchmark~\cite{wu2024longmemeval}. RMM~\cite{tan2025prospect} effectively transcends the limitations of traditional fixed‑length chunk boundaries through topic‑driven dynamic summarization. Beyond these plain index-based approaches, Mem0‑g~\cite{chhikara2025mem0} represents memory as a labeled graph where entities act as nodes and relations act as edges. SG‑mem~\cite{wu2025sgmem} departs from prior practices of constructing knowledge graphs by building graphs solely from texts and sentences, which achieves superior performance. Zep and CAM~\cite{rasmussen2025zep,li2025cam} introduce a label propagation algorithm to mitigate the limitations of traditional GraphRAG~\cite{edge2024local} in handling incremental updates of communities.

\subsection{Difference between RAG and Memory}
RAG and agent memory systems, despite their separate origins, now employ substantially overlapping techniques~\cite{hu2025memory}. RAG was initially conceived to connect LLMs to static knowledge sources, whereas agent memory prioritized knowledge updating and persistence. As both fields advance, their methodologies increasingly overlap. For instance, works like HippoRAG~\cite{jimenez2024hipporag,gutierrez2025rag} are now recognized in both RAG and memory research as solutions to the long-term memory challenges of large language models.

Empirically, RAG tasks typically build indexes directly over raw content such as documents and do not require frequent updates, whereas dialog memory systems tend to index extracted “memories” from conversations and place greater emphasis on online index maintenance.
Drawing on both research traditions, we propose a unified framework tailored for long-term dialogue memory.

\section{A Unified Memory Framework} \label{sec:unified}

\begin{table*}[htbp]
\centering
\setlength{\tabcolsep}{4pt}
\scriptsize
\begin{tabular}{lllllll}
\toprule
 & \textbf{Key} & \textbf{Value} & \textbf{Query} & \textbf{Index (Struct; Op)} & \textbf{Retrieval} & \textbf{Answering} \\
\midrule
\textbf{LongMemEval} 
& Session + Facts 
& Session 
& Q + T 
& F; Add 
& Q $\rightarrow$ K $\rightarrow$ V 
& CoN \\

\textbf{RMM} 
& Topic summ 
& Session + Key
& Q
& F; Add, Upd 
& Q $\rightarrow$ K $\rightarrow$ V $\rightarrow$ Rerank 
& Direct \\

\textbf{A-Mem} 
& Session + Kw, Tag, Summ
& Key 
& Q 
& G; Add, Upd 
& Q $\rightarrow$ K $\rightarrow$ V 
& Direct \\

\textbf{Mem0-G} 
& Entity name, Triple 
& Triple 
& Q
& G; Add, Align 
& Q $\rightarrow$ K $\rightarrow$ 1-Hop $\rightarrow$ V
& Direct \\

\textbf{Zep} 
& Entity summ, Triple, Comm 
& Key 
& Q 
& HG; Add, Align, Comm Upd 
& Q $\rightarrow$ K $\rightarrow$ 1-Hop $\rightarrow$ V $\rightarrow$ Rerank
& Direct \\
\bottomrule
\end{tabular}
\caption{Decomposition of some representative dialog memory systems under the unified framework. A more complete version and detailed explanations can be found in Appendix \ref{appendix_relatedsystem}.}
\label{tab:memory_comparison}
\end{table*}

\begin{figure*}[!ht]
  \includegraphics[width=\textwidth]{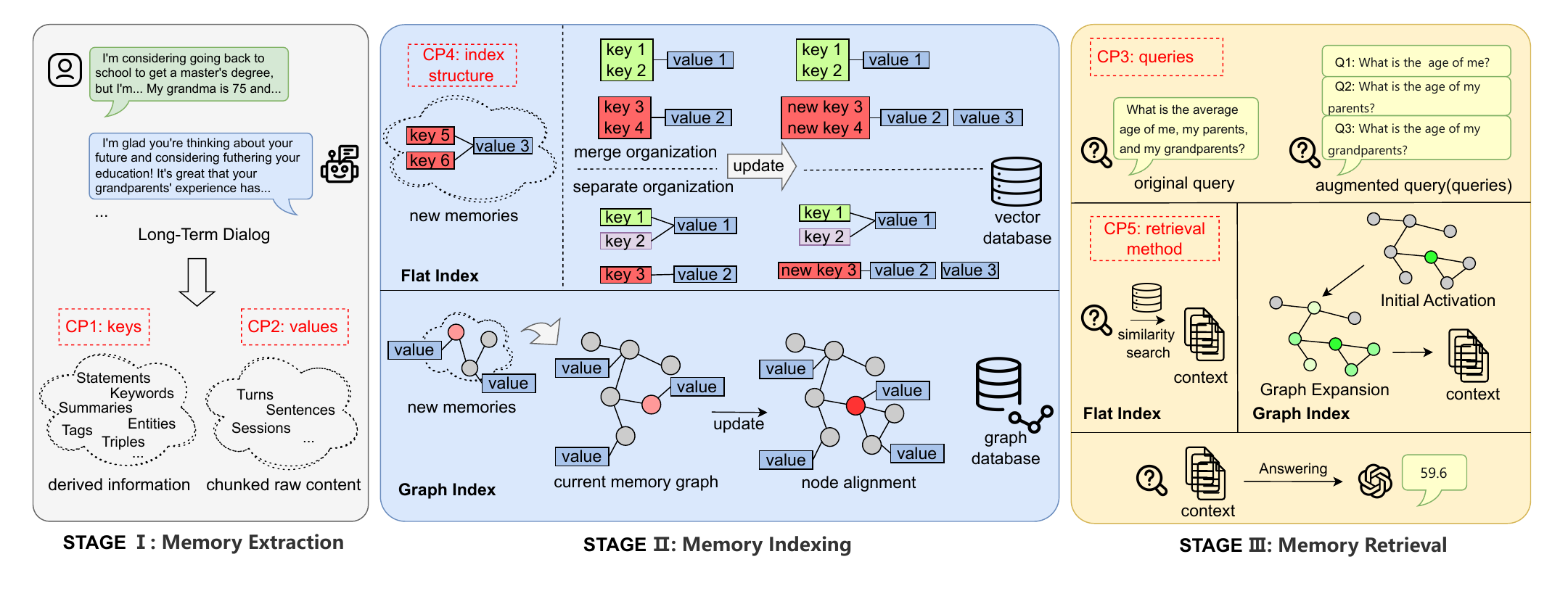}
  \caption{The unified framework for dialog memory system.}
  \label{fig:unified}
  \vspace{-0.1in}
\end{figure*}

We abstract dialogue memory and RAG-style memory systems into a six-tuple:
\[
  \langle K, V, Q, I, R, A \rangle ,
\]
where $K$ denotes the set of \emph{keys} (memory units), $V$ denotes the \emph{values} (evidence used in the answering stage $A$), $Q$ denotes the \emph{queries} (retrieval requests), $I$ denotes the \emph{index structure} and $R$ denotes the \emph{retrieval method}.

The overall pipeline can be decomposed into four stages:
\begin{compactitem}
  \item \textbf{Memory Extraction:} extract memory from raw dialogue, i.e., generate keys and values;
  \item \textbf{Memory Indexing:} organize keys into a flat index or graph structure, building and maintaining $I$;
  \item \textbf{Memory Retrieval:} given a query $q \in Q$, get target keys in $K$ using retrieval method $R$, map back to the corresponding values in $V$.
  \item \textbf{Question Answering:} assemble the final context and answer the user input.
\end{compactitem}

In practical applications, the above four operational stages must be invoked repeatedly as the dialog progresses. The invocation timing depends on the specific strategy. For example, memory extraction and indexing may be performed at the end of each session, while memory retrieval and answering may be executed after each user input.

Table~\ref{tab:memory_comparison} shows how representative dialog memory systems can be decomposed and aligned under this unified framework, enabling comparison across otherwise heterogeneous implementations.
Appendix \ref{appendix:unified_framework} provides additional details for each stage.

\subsection{Keys and Values} \label{sec:key_value}

Keys usually represent important information derived from dialog history, which can be seen as memory units.
Depending on the design, keys may correspond to summaries, factual statements, keywords~\cite{xu2025mem}, or entities and triples~\cite{rasmussen2025zep}, and may encode either semantic or episodic information~\cite{du2025rethinking}.
As illustrated in Table~\ref{tab:memory_comparison}, some existing work also considers raw content (e.g., sessions) as one form of Key~\cite{wu2024longmemeval}.

Values correspond to the evidence ultimately provided to the answering model.
Using the retrieved keys as values is a natural choice, and some prior work~\cite{tan2025prospect} additionally (or alternatively) uses the raw content as values.
Raw content such as sessions preserve contextual continuity but incur higher storage and reasoning costs, whereas derived information are usually compressed, which improve efficiency at the risk of information loss.

\subsection{Query and Answering} \label{sec:query}
Query Augmentation has been widely adopted in RAG systems~\cite{guo2024lightrag, gutierrez2025rag}, to improve the understanding of user intent during the retrieval stage and to enhance the recall quality of relevant documents. 
Common operations include query rewriting, expansion, and decomposition, among others.
This paradigm has recently been extended to dialogue memory tasks in several studies~\cite{zhang2025personalize}.
The answer generation stage can also benefit from several optimization techniques, such as extracting key information before answering \cite{yu2311enhancing}.

Since query augmentation and answering optimization are relatively independent, they are not discussed in detail in this work, and the original input is used as the query by default.

\subsection{Index Structure} \label{sec:index}

The index structure specifies how keys are organized and maintained, providing the substrate on which retrieval operates.

\subsubsection{Flat Index and Key Organization}

In the simplest designs, each key is stored as an independent vector entry under the \textbf{Separate Organization}. The separate organization results in a large number of vectors and suffers from issues of information fragmentation and redundancy.
While some keys (usually derived from same session) can be concatenated to jointly produce a single vector representation with the \textbf{Merge Organization}.
More specifically, A-Mem~\cite{xu2025mem} introduces the concept of Memory Notes and aggregate all keys derived from the same value into a single vector (which we refer to as the Merge-by-session). While LongMemEval further explores aggregating only keys of the same type into one vector (which we refer to as the Merge-by-type).

\subsubsection{Graph Index}

Graph indices connect keys via explicit edges.
In graph-based designs, nodes may represent textual units or entities, while edges capture relationships such as semantic relatedness, structural association, or temporal order.
Some systems further adopt hierarchical graph structures to connect multiple abstraction levels (e.g., entity--community hierarchies~\cite{edge2024local}).

\subsubsection{Index Maintenance and Update}

In theory, memories acquired at different times may exhibit relationships such as duplication, completion, or update and replacement. Common memory maintenance operations~\cite{chhikara2025mem0} include \textit{Add}, \textit{Update}, \textit{Delete} and \textit{Noop}. 
Recent work~\cite{ong2025towards} argues that directly deleting old memories is undesirable, as they may still be useful in the future (e.g., when a user refers to a former occupation).

In practice, these operations are generally applicable to both flat index and graph-based index. Figure \ref{fig:unified} Stage II illustrates the \textit{Update} operation. In graph-based settings, the \textit{Update} operation is typically referred to as node merge (also known as entity alignment) and the corresponding edge merge (relation alignment).

For hierarchical graph indices, additional maintenance operations are required, such as community reconstruction~\cite{edge2024local} or incremental updates~\cite{rasmussen2025zep,li2025cam}.

\subsection{Retrieval Procedure} \label{sec:retrieval}

\subsubsection{Retrieval over Flat Index}

For flat indices, retrieval typically follows a standard vector-search pipeline: embed the query $q$ into a vector representation, then perform approximate or exact nearest-neighbor search over the key embeddings, and select top-$k$ keys according to similarity scores.
Finally map these keys back to their associated values and assemble the context.

Note that keys and values are often not in a one-to-one relationship. Therefore, to ultimately obtain the top-$n$ values for answering a query, the value of $k$ is typically set larger than $n$, and a value-level re-ranking mechanism is required.

\subsubsection{Retrieval over Graph Index} \label{method:graph_retrieval}
Graph-based retrieval is defined as retrieving a subgraph (or some nodes) conditioned on a query $q$. 
The graph retrieval process typically combines an initial activation step with graph-based expansion.

\paragraph{Initial Activation}
Similar to flat indices, first obtain an initial set of seed nodes: embed $q$ and perform similarity search over node embeddings, select top-$k$ nodes. Or allow direct vector matching between $Q$ and triples~\cite{chhikara2025mem0}.

\paragraph{Graph Expansion}
Starting from the seed nodes, graph expansion methods aim to gather additional context from the graph. A common approach is BFS-style local expansion~\cite{rasmussen2025zep}, which collects nodes within a limited hop distance (typically one hop) and truncates the result when a predefined maximum number of nodes is reached. Beyond this, more advanced structure-aware~\cite{gutierrez2025rag} or semantics-aware~\cite{zhang2025bridging} methods guide expansion using graph topology or embedding similarity.

\section{Experiments}
To understand how different design choices and underlying implementation affect the behavior of dialog memory systems, we conduct a series of controlled, stage-wise experiments under the unified framework introduced in Section~\ref{sec:unified}.

The experiments are organized as follows.
Sections~\ref{exp:key} and~\ref{exp:operation} examine flat-based methods and aim to identify strong baseline configurations.
Sections~\ref{exp:build} and~\ref{exp:search} focus on graph-based methods, analyzing graph construction strategies and retrieval mechanisms.
Section~\ref{exp:end2end} presents the end-to-end experimental results based on selected settings.

Due to space limit, we report only a subset of the results in the main paper; more detailed experimental results (including efficiency analysis) can be found in the Appendix \ref{appendix_extendedexperimentalresults}.

\subsection{Datasets and Settings} \label{exp:setting}

We conduct experiments on two representative benchmarks, LongMemEval and HaluMem, which are designed to evaluate long-term dialog memory from complementary perspectives.

\textbf{LongMemEval}~\cite{wu2024longmemeval} primarily focuses on memory retrieval and reasoning over extremely long dialog histories and consists of two subsets (S and M). 

\textbf{HaluMem}~\cite{chen2025halumem} is designed to evaluate memory extraction, updating, and consistency. HaluMem contain a large number of information updates and explicitly test whether memory systems can correctly extract new information, update outdated memories, and avoid hallucinated responses. It consists of Medium and Long subsets, due to the high computational cost, we conduct experiments only on the Medium subset.

By default, we use the same experimental setup across all experiments: LLaMA-3.1-8B is used as both the memory extraction (graph construction) model and the answer generation model, while Contriever~\cite{izacard2021unsupervised} is used as the embedding model.
We use gpt-4o as the judge model for LongMemEval and use gpt-4o-mini as the judge model for HaluMem.
During the QA experiments, for Value = session, we select the top-$5$ values to generate the answer, whereas for Value = Key, we select the top-$20$ values for answer generation.
All experiments are conducted on a machine with a machine with 8 NVIDIA H100 80GB GPUs. 

\subsection{How to select Keys} \label{exp:key}

\begin{table}[t]
\centering
\small
\setlength{\tabcolsep}{4pt}
\begin{tabular}{lcccc}
\toprule
\multirow{2}{*}{Key Design} &
\multicolumn{2}{c}{LME-S} &
\multicolumn{2}{c}{LME-M} \\
\cmidrule(lr){2-3}\cmidrule(lr){4-5}
& R@5 & R@10 & R@5 & R@10 \\
\midrule
\texttt{session} & 0.9021 & 0.9714 & 0.7112 & 0.8043 \\
\texttt{session,S,F,K} & \textbf{0.9379} & 0.9690 & 0.7327 & 0.8521 \\
\texttt{[session,S,F,K]} & 0.9165 & 0.9666 & 0.7184 & 0.8210 \\
\texttt{session,[S,F,K]} & \textbf{0.9379} & \textbf{0.9833} & \textbf{0.7685} & \textbf{0.8592} \\
\midrule
\texttt{S,F,K} & \textbf{0.9117} & 0.9618 & 0.6921 & 0.8091 \\
\texttt{[S,F,K]} & 0.9045 & \textbf{0.9642} & \textbf{0.7064} & \textbf{0.8138} \\
\bottomrule
\end{tabular}
\caption{Retrieval performance on LongMemEval under different keys. S=summary, F=aggregated factual statements, K=aggregated keywords. \texttt{[a,b]} means their representations are merged during retrieval. R@n means Recall@n.}
\label{tab:key_ablation}
\vspace{-0.2in}
\end{table}

\begin{table*}[t]
\centering
\small
\setlength{\tabcolsep}{5pt}
\begin{tabular}{lccccccc}
\toprule
\multirow{2}{*}{Key Design} &
\multicolumn{3}{c}{op: add} &
\multicolumn{4}{c}{op: add / update / noop} \\
\cmidrule(lr){2-4}\cmidrule(lr){5-8}
& Mem-R & Mem-P & QA-C
& Mem-R & Mem-P & MemUpdate-C & QA-C \\
\midrule
\texttt{S,F,K} & 0.7332 & 0.9360 & 0.2815 & 0.8069 &  0.8829 & 0.1416 & 0.5134 \\
\texttt{[S,F,K]} & 0.7332 & 0.9360 & \textbf{0.4785} & 0.8057 & 0.8846 & \textbf{0.2207} & \textbf{0.5861} \\
\bottomrule
\end{tabular}
\caption{Performance on HaluMem-medium under different operation settings.
“Mem-R” denotes Memory Extraction Recall, “Mem-P” denotes Target Memory Extraction Precision, MemUpdate-C denotes Correct Ratio of memory update operation, and QA-C denotes Correct Ratio of end-to-end Question Answering.}
\label{tab:op_ablation}
\end{table*}

As discussed in Section \ref{sec:key_value}, existing works differ substantially in the representation forms they adopt in practice.
The design of keys plays a critical role in memory systems. This involves two main questions:
(1) what types of keys to extract, and
(2) how to organize these keys into memory.

In this section, we analyze and evaluate three commonly used types of keys—\textit{summaries}, \textit{factual statements}, and \textit{keywords}—together with two organization strategies,\textit{ merge-by-type} and \textit{merge-by-session}~\footnote{We omit the separate strategy as it has already been shown to perform poorly in the LongMemEval paper. Interestingly, we find that the separate strategy achieves the best performance on the HaluMem dataset instead; details can be found in the Appendix \ref{appendix_extendedexperimentalresults}.}.
For \textit{merge-by-type}, the query is matched against each merged content to obtain similarity scores, which are averaged to produce the final score for the session.

As shown in Table \ref{tab:key_ablation}, augmenting session-based keys (i.e., raw dialogue content) with three types of derived information at different granularities consistently improves retrieval performance, regardless of the organization strategy employed.
In contrast, the merge-by-type organization (\texttt{session,S,F,K}) outperforms merge-by-value (\texttt{[session,S,F,K]}), which differs from the conclusions reported in LongMemEval. 
Moreover, we find that merging only the derived information while keeping the raw content as an independent key (\texttt{session,[S,F,K]}) yields better performance than the standard merge-by-value strategy.
For settings where raw content is not allowed as a key, the merge-by-value strategy shows a slight advantage.

\paragraph{Takeaway.}
If retaining the session is allowed, use \texttt{session,[S,F,K]} as keys organization, otherwise try (\texttt{[S,F,K]}) first.

\subsection{How to select Memory Operations} \label{exp:operation}

As discussed in Section~\ref{sec:index}, a key distinction between dialog memory systems and standard RAG pipelines is that memory indexing often involves explicit maintenance operations beyond simple insertion.
Although many existing memory systems support update or alignment mechanisms in principle, their reported end-to-end results are often outperformed by systems that only perform add operations\footnote{Details can be found in Table \ref{tab:memory_misalignment}.}.
As a result, the impact of different maintenance operations is rarely isolated or systematically analyzed.

In this section, we explicitly evaluate the effectiveness of \textit{Update} and \textit{Noop} operations on HaluMem, a benchmark designed to assess memory extraction and update.
Following recent findings~\cite{ong2025towards} that direct deletion is generally unnecessary in dialog memory systems, we omit the \textit{Delete} operation from our analysis.

As shown in Table \ref{tab:op_ablation}, introducing the \textit{Update} and \textit{Noop} operations leads to a decrease in memory precision, likely because these operations may incorrectly modify memories. In contrast, memory recall improves substantially, which ultimately translates into higher end-to-end QA accuracy.
Comparing the \textit{merge-by-type} and \textit{merge-by-value} key organization strategies, we observe that latter consistently performs better, regardless of whether the \textit{Update} and \textit{Noop} operations are applied. This results are consistent with Table~\ref{tab:key_ablation} and exhibit more pronounced differences, which may be attributed to the characteristics of the dataset.

\paragraph{Takeaway.}
The \textit{Update} and \textit{Noop} operations are indeed useful and are worth considering in practice.

\subsection{How to construct graphs} \label{exp:build}

\begin{table}[t]
\centering
\small
\setlength{\tabcolsep}{6pt}
\begin{tabular}{lcccc}
\toprule
\multirow{2}{*}{Graph Design} &
\multicolumn{2}{c}{LME-S} &
\multicolumn{2}{c}{LME-M} \\
\cmidrule(lr){2-3}\cmidrule(lr){4-5}
& R@5 & R@10 & R@5 & R@10 \\
\midrule
\texttt{[S,F,K]} & 0.9045 & 0.9642 & 0.7064 & 0.8138 \\
SimGraph & 0.8424 & 0.9498 & 0.6043 & 0.7766 \\
\midrule
KnowGraph &  0.9116 & 0.9665  &  0.6610 & 0.7828 \\
DescGraph & \textbf{0.9331} & \textbf{0.9713} & \textbf{0.7661} & \textbf{0.8735} \\
\bottomrule
\end{tabular}
\caption{Retrieval performance on LongMemEval under different graph schema.}
\label{tab:build_ablation}
\end{table}

As discussed in Section \ref{sec:index}, the key to constructing a graph index lies in how nodes and edges (connections) are defined.
For lightweight design, many existing works~\cite{xu2025mem,wu2025sgmem} directly build upon an existing flat index, introducing edges based on textual similarity or membership relations.
Another line of work~\cite{jimenez2024hipporag,chhikara2025mem0} leverages LLMs to extract entities and semantic relations from raw content, thereby constructing a knowledge-graph-style memory index.

In this section, based on LongMemEval, we compare these two common graph construction paradigms and further propose a simple yet effective improvement.

\textbf{SimGraph}: Following A-mem, for each Key Group $[S_i, F_i, K_i]$, we retrieve the top-$5$ most similar Key Groups via vector matching and then employ an LLM to determine whether an edge should be established.

\textbf{KnowGraph}: A Knowledge Graph composed of triples \(\langle subject, predicate, object \rangle\), where nodes (subjects/objects) correspond to entities and edges (predicates) represent relations between entities.
A potential limitation of this construction paradigm is its restricted representational capacity. 
Similar to factual statements, triples are primarily suited for expressing semantic memory, but are less effective at capturing episodic memory. In flat indices, episodic memory is typically carried by summaries, which allow agents to describe and record events in a relatively free-form manner.

\textbf{DescGraph}: We adopt a direct approach by enriching entities with natural language descriptions, which record both semantic memory and episodic memory associated with each entity. During the construction of DescGraph, these descriptions are updated as nodes and edges are merged, in a manner analogous to the memory update operation in flat indices.

\textbf{Retrieval Setup}: For all graph variants, we employ the same retrieval strategy: Q $\rightarrow$ K $\rightarrow$ 1-Hop $\rightarrow$ V (first match initial nodes and then perform a one-hop expansion, values are selected by ranking nodes according to their query–node similarity scores.).
For KnowGraph, node representations are derived from entity names, whereas for DescGraph, node representations are based on node descriptions.

As shown in Table \ref{tab:build_ablation}, using SimGraph actually led to worse retrieval performance. The underlying reason is that graph expansion based on similarity edges introduces excessive noise without an effective reranking mechanism, making it less effective than using only the initially activated nodes.
The experimental results demonstrate that DescGraph consistently achieves superior performance. More details can be found in Appendix \ref{appendix_casestudy}. 

\paragraph{Takeaway.}
Constructing graphs based on similarity is hard to make effective, whereas adding entity descriptions on top of an entity–relation–based graph yields clear improvements.

\subsection{How to search graphs} \label{exp:search}

\begin{table}[t]
\centering
\small
\setlength{\tabcolsep}{6pt}
\begin{tabular}{lcccc}
\toprule
Activation Design & R@5 & R@10 & N@5 & N@10 \\
\midrule
Entities & \textbf{0.9642} & \textbf{0.9905} &  \textbf{0.9599} & \textbf{0.9648} \\
Triples & 0.9403 & 0.9809 & 0.9363 & 0.9460 \\
\bottomrule
\end{tabular}
\caption{Graph Activation method comparison on LongMemEval-S. N@n means NDCG@n.}
\label{tab:activate_ablation}
\end{table}

\begin{table}[t]
\centering
\small
\setlength{\tabcolsep}{6pt}
\begin{tabular}{lcccc}
\toprule
\multirow{2}{*}{Re-rank Design} &
\multicolumn{2}{c}{no expansion} &
\multicolumn{2}{c}{one-hot expand} \\
\cmidrule(lr){2-3}\cmidrule(lr){4-5}
& R@5 & R@10 & R@5 & R@10 \\
\midrule
$Score_s$ & 0.9546 & 0.9904 & 0.8615 & 0.9618 \\
\midrule
$Score_e$ & \textbf{0.9642} & 0.9904 & 0.9594 &  0.9643 \\
$(Score_e, Score_g)$ & \textbf{0.9642} & 0.9904 & \textbf{0.9642} & \textbf{0.9928} \\
\bottomrule
\end{tabular}
\caption{Retrieval performance on LongMemEval under different ranking methods.}
\label{tab:rank_ablation}
\end{table}

\begin{table*}[t]
\centering
\small
\begin{tabular}{l l l l c c c c c c}
\toprule
\multirow{2}{*}{Extraction} &
\multirow{2}{*}{Embedding} &
\multirow{2}{*}{Answering} &
\multirow{2}{*}{Index} &
\multicolumn{4}{c}{Retrieval} &
\multicolumn{2}{c}{Answer Accuracy} \\
\cmidrule(lr){5-8} \cmidrule(lr){9-10}
& & & &
R@5 & R@10 & N@5 & N@10 &
V=session & V=Key \\
\midrule

\rowcolor{gray!15}
\multicolumn{10}{c}{\textbf{LongMemEval-S}} \\
\midrule
llama-3.1-8b & contriever & llama-3.1-8b & flat
  & 0.9045 & 0.9642 & 0.9100 & 0.9207 & 0.614 & \textbf{0.570} \\
llama-3.1-8b & contriever & llama-3.1-8b & graph
  & \textbf{0.9356} & \textbf{0.9761} & \textbf{0.9393} & \textbf{0.9477} & \textbf{0.620} & 0.518 \\
gpt-4o-mini & text-emb-3-s & gpt-4o & flat
  & 0.9284 & 0.9880 & 0.9346 & 0.9458 & 0.760 & \textbf{0.752} \\
gpt-4o-mini & text-emb-3-s & gpt-4o & graph
  & \textbf{0.9690} & \textbf{0.9928} & \textbf{0.9711} & \textbf{0.9742} & \textbf{0.892} & 0.690 \\
  
\midrule
\rowcolor{gray!15}
\multicolumn{10}{c}{\textbf{LongMemEval-M}} \\
\midrule
llama-3.1-8b & contriever & llama-3.1-8b & flat
  & 0.7064 & 0.8138 & 0.7309 & 0.7600 & 0.526 & \textbf{0.478} \\
llama-3.1-8b & contriever & llama-3.1-8b & graph
  & \textbf{0.7661} & \textbf{0.8735} & \textbf{0.8138} & \textbf{0.8377} & \textbf{0.548} & 0.428 \\
gpt-4o-mini & text-emb-3-s & gpt-4o & flat
  & 0.7231 & 0.8353 & 0.7292 & 0.7588 & 0.638 & \textbf{0.620} \\
gpt-4o-mini & text-emb-3-s & gpt-4o & graph
  & \textbf{0.8281} & \textbf{0.9307} & \textbf{0.8631} & \textbf{0.8880} & \textbf{0.754} & 0.592 \\
  
\bottomrule
\end{tabular}
\caption{Comparison of two strong baselines on LongMemEval. flat: Key=\texttt{[S,F,K]}, Op=Add. graph: Key=entity description, Retrieval=Q $\rightarrow$ K $\rightarrow$ V, Rank=$(Score_e, Score_g)$}
\label{tab:main_results_lme}
\end{table*}

\begin{table*}[t]
\centering
\small
\begin{tabular}{l l l l c c c c c c}
\toprule
\multirow{2}{*}{Extraction} &
\multirow{2}{*}{Embedding} &
\multirow{2}{*}{Answering} &
\multirow{2}{*}{Index} &
\multicolumn{3}{c}{Extraction} &
\multicolumn{3}{c}{Answer (V=Key)} \\
\cmidrule(lr){5-7} \cmidrule(lr){8-10}
& & & &
Mem-R & Mem-P & MemUpdate-C &
QA-C & QA-H & QA-O \\
\midrule
llama-3.1-8b & contriever & llama-3.1-8b & flat
  & \textbf{0.8057} & 0.8846 & 0.2207 & \textbf{0.5861} & \textbf{0.1855} & \textbf{0.2284 }\\
llama-3.1-8b & contriever & llama-3.1-8b & graph
  & 0.4742 & \textbf{0.9921} & - & 0.4935 & 0.2307 & 0.2757 \\
gpt-4o-mini & text-emb-3-s & gpt-4o & flat
  & \textbf{0.7759} & 0.8809 & 0.2691 & 0.5645 & \textbf{0.1578} & 0.2778 \\
gpt-4o-mini & text-emb-3-s & gpt-4o & graph
  & 0.6493 & \textbf{0.9936} & - & \textbf{0.6322} & 0.2059 & \textbf{0.1618} \\

\bottomrule
\end{tabular}
\caption{Comparison of two strong baselines on HaluMem-medium. flat: Key=\texttt{[S,F,K]}, Op=Add/Update/Noop. graph: Key=entity description, Retrieval=Q $\rightarrow$ K $\rightarrow$ V, Rank=$(Score_e, Score_g)$}
\label{tab:main_results_halumem}
\vspace{-0.1in}
\end{table*}

As introduced in Section \ref{method:graph_retrieval}, graph retrieval can be decomposed into two stages: activation and expansion.
In this section, we compare two commonly used graph activation strategies:
\begin{compactitem}
\item \textit{Entity activation}: the query is matched against entity embeddings, and the top-$k$ entities are selected.
\item \textit{Triple/Relation activation}: the query is matched against triple embeddings, and the top-$k$ triples are selected.
\end{compactitem}
We further evaluate one of the most commonly adopted graph expansion methods, 1-hop expansion: traverses each initially activated entity and add its 1-hop neighbors to the activated entity set.

Since the number of values (i.e. sessions) associated with the activated entity set may exceed the return budget of the system, a re-ranking step is required. We consider two straightforward scoring strategies:
\begin{compactitem}
\item $Score_s$: the vector similarity between the query and the session (value).
\item $Score_e$: the vector similarity between the query and the entity (key). 
\end{compactitem}
As a single entity may correspond to multiple sessions, relying solely on $Score_e$ often leads to score ties. We therefore introduce an additional secondary order key $Score_g$ denoting the number of candidate keys associated with a value, where candidate keys include the currently activated entities and their 1-hop neighbors.

The two methods reported in Table \ref{tab:activate_ablation} do not apply graph expansion and use the same re-ranking strategy ($S_e, S_g$). The results show that direct entity activation consistently outperforms triple-based activation.
Furthermore, Table \ref{tab:rank_ablation} indicates that, under the augmented re-ranking configuration, the performance difference between enabling and disabling 1-hop expansion is marginal.

However, when only $Score_s$ is used, the performance of the expansion degrades noticeably. This is because 1-hop expansion introduces a larger set of candidate values, thereby imposing a higher demand on the re-ranking quality. After incorporating $Score_g$ as a secondary key to $Score_e$, the relative improvement under the 1-hop expansion setting becomes more pronounced, further supporting this explanation.

\paragraph{Takeaway.}
If Value=session, directly activating entities (w/o expansion) and rank value according to $(Score_e, Score_g)$ is a strong baseline.

\subsection{End-to-end Results} \label{exp:end2end}

Finally, we adopt a setting that disallows using raw content as keys and select one best-performing configuration for each of the flat based and graph based methods on both LongMemEval and HaluMem, and conduct a full comparison of their results.
More results can be found in Appendix \ref{appendix_extendedexperimentalresults}.

\textbf{LongMemEval.}
In terms of retrieval performance, the graph method consistently outperforms the flat index method across different model combinations. This gap becomes more pronounced as the dataset scale increases from S to M.

For question answering, under the Value = session setting, the graph method achieve better performance overall, with particularly notable improvements under the higher-capacity setting (using gpt-4o-mini for graph construction). This suggests that, compared to directly extracting flat key such as factual statements, \textit{graph construction (extracting entities, relations, and descriptions) requires stronger model capabilities}.

Under the Value=Key setting, however, the graph method performs worse. This is because each key representation used in the flat index method (i.e., \texttt{[S,F,K]}) aggregates all derived information from a single session. In contrast, the information contained in entity descriptions is not sufficiently rich. 
Consequently, when the same number of keys is selected for answering, the flat index method effectively has access to substantially more information—essentially comparable to using sessions.

\textbf{HaluMem.}
HaluMem does not include explicit evaluation of retrieval metrics; instead, it measure the memory extraction performance.
The QA results on HaluMem are consistent with the conclusions drawn on LongMemEval: when using a weaker extraction model and setting V=key, the graph method performs poorly.
Notably, graph method also exhibit lower recall in memory extraction metrics, which may stem from the granularity mismatch between the ground-truth memories (biased toward factual statements) and the memories represented in graphs (entity-centric descriptions).

\section{Conclusion}

This paper presents a systematic analysis of long-term dialog memory architectures.
Motivated by inconsistent empirical findings in prior work, we propose a unified analytical framework that decomposes dialog memory systems into core components, enabling controlled comparison across graph-based and non-graph approaches.

We evaluate common design choices in dialog memory systems, 
Our results show that (1) \textbf{foundational system settings have a substantial impact on performance}; (2) \textbf{while graph-based memory can offer advantages under certain configurations, inappropriate graph construction or retrieval strategies may instead degrade results}.

Overall, this work provides a practical analytical framework and a set of strong baselines that help clarify the design space of dialog memory systems.

\section*{Limitations}

First, the unified memory framework we propose is intended to cover the majority of commonly used memory systems, helping the community focus on shared underlying implementation details and facilitating fair comparisons. However, we acknowledge that the proposed framework cannot cover all possible memory systems. In practice, memory systems can be highly customized; for example, Memory OS treats memory as a schedulable system resource and manages it by partitioning memory into long-term, mid-term, and short-term components.

Second, to derive relatively general conclusions and a set of strong baselines, we conduct experimental analyses under several commonly adopted settings at different stages of the pipeline. We do not exhaustively explore all possible design choices, such as alternative forms of memory keys, heterogeneous or hierarchical graphs~\cite{gutierrez2025rag}, more sophisticated graph retrieval methods~\cite{zhuang2025linearrag,luo2025kbqa,8085196,hu-etal-2018-state}, or approaches that require additional training~\cite{wang2025mem,yan2025memory}. Meanwhile, more fine-grained comparative experiments remain to be explored, such as disentangling and comparing the Update and Noop operations.

Third, our experiments are conducted on two widely recognized and complementary benchmarks, LongMemEval and HaluMem. Nevertheless, some of our conclusions may vary on other memory benchmarks with different characteristics. In fact, selecting different configurations for different datasets (and real-world application scenarios) is a reasonable and practical choice.

Finally, considering both local deployment and API-based settings, we only evaluate two backbone model configurations—LLaMA-3.1-8B with Contriever, and OpenAI GPT-4o-mini / GPT-4o with text-3-embedding-small—and do not explore a broader range of backbone model combinations.


\bibliography{custom}

\clearpage

\appendix
\section*{Appendices}

\counterwithin{figure}{section}
\counterwithin{table}{section}
\section{Unified Framework Details}
\label{appendix:unified_framework}

\subsection{Forms of Keys}
Common forms of Keys include:
\begin{compactitem}
  \item \textbf{Summaries:} typically in session-level;
  \item \textbf{Statements:} natural language assertions, suitable for factual content or event description;
  \item \textbf{Keywords:} key phrases extracted from the original text;
  \item \textbf{Tags/Categories:} abstract labels for grouping or classification;
  \item \textbf{Entities:} persons, organizations, products, concepts, etc.;
  \item \textbf{Triples:} \(\langle subject, predicate, object \rangle\), typically used with graph structures.
\end{compactitem}

\subsection{Semantic or Episodic Memory}

Inspired by human memory, recent work tends to categorize dialogue memory into:
\begin{compactitem}
  \item \textbf{Semantic Memory:} stable facts, attributes, and relationships; often represented as statements or triples (e.g., “the user likes sugar-free drinks” or \(\langle \text{User}, \text{likes}, \text{sugar-free drinks} \rangle\));
  \item \textbf{Episodic Memory:} event-like records of “who did what, where, and when”; represented as summaries or statements with time, location, and participating entities (e.g., “On 2024-11-10 in Shanghai, the user asked about EU privacy-law compliance.”).
\end{compactitem}

\subsection{Memory Index Operations}

\begin{compactitem}
    \item \textbf{Add}: Newly extracted memories are appended to the index as new entries.
    \item \textbf{Update}: Certain existing memories are updated or revised based on newly extracted ones.
    \item \textbf{Delete}: Some existing memories are removed from the index.
    \item \textbf{Noop}: Newly extracted memories are discarded and not added to the index.
\end{compactitem}

Recent work~\cite{ong2025towards} argues that directly deleting old memories is undesirable, as they may still be useful in the future (e.g., when a user refers to a former occupation). When a user query only concerns the most recent information, outdated memories can instead be filtered out during retrieval using timestamps.

\subsection{Graph Index and Retrieval}

Graph indices connect keys via explicit edges.

\paragraph{Edge Types}

Common edge types include:
\begin{compactitem}
  \item \textbf{Similarity edges:} connect semantically similar keys according to embedding distance;
  \item \textbf{Part-of edges:} represent that a key belongs to a higher-level node;
  \item \textbf{KG-style relation edges:} as in HippoRAG2 and Mem0, edges are labeled with predicates between entities.
  \item \textbf{Temporal edges:} capture temporal order (e.g., adjacency within the same session);
  \item \textbf{Causal edges:} represent causal relationships where one event leads to another;
\end{compactitem}

\paragraph{Node Types}

Different systems instantiate graph nodes in different ways:
\begin{compactitem}
  \item \textbf{Text chunks as nodes:} e.g., CAM, where chunk-level text segments are nodes;
  \item \textbf{Sentences as nodes:} e.g., SG-mem, using sentence-level nodes for finer granularity and more precise selection;
  \item \textbf{Entities as nodes:} e.g., GraphRAG, Zep, Mem0:
  \item \textbf{Hybrid nodes:} e.g., HippoRAG2 uses a hybrid scheme that can be viewed as hierarchical:
\end{compactitem}

\paragraph{Hierarchical Graphs}

Hierarchical graphs connect multiple semantic levels:
Some prior work, such as SGMem, treats the inherent subordinate relationship between keys and values as “edges,” thereby constructing a so-called “hierarchical graph.” In contrast, in our framework this relationship is regarded as a fixed mapping. Even a flat index can naturally preserve such a mapping, and therefore it is not considered a hierarchical graph.

Correspondingly, more representative examples of true hierarchical graphs include the Insight–Query–Interaction hierarchy in G-Memory and the Community–Entity hierarchical structure in GraphRAG. Since the former is primarily designed for system memory, this paper focuses its discussion on the latter.

\paragraph{Graph Expansion}

Starting from the seed nodes, expand over the graph to gather additional context:
\begin{compactitem}
  \item \textbf{Simple neighbor expansion (BFS-style):} take $h$-hop neighbors (often $h=1$) and truncate to a maximum number of nodes $N_{\max}$ when necessary;
  \item \textbf{Structure-aware expansion (HippoRAG2):} run PageRank or related algorithms from the seed nodes, exploiting graph topology; node embeddings are used mainly to re-rank the expanded set;
  \item \textbf{Semantic-aware expansion (EcphoryRAG):} Compute a ``center vector'' from the seed nodes, and teratively expand by selecting top-$k$ nodes semantically close to the center vector. Finally re-score all candidates against the query embedding to avoid semantic drift.
\end{compactitem}

\section{Dataset Details}

\textbf{LongMemEval}~\cite{wu2024longmemeval} primarily focuses on memory retrieval and reasoning over extremely long dialog histories and consists of two subsets (S and M). The benchmark covers a diverse set of task types, including information extraction, cross-session reasoning, temporal reasoning, knowledge updating, and refusal detection, making it well suited for evaluating memory organization and retrieval quality at scale.

\textbf{HaluMem}~\cite{chen2025halumem} is designed to evaluate memory extraction, updating, and consistency. HaluMem dialogs contain a large number of information updates and explicitly test whether memory systems can correctly extract new information, update outdated memories, and avoid hallucinated responses. Different from LongMemEval, HaluMem does not include retrieval-based evaluation metrics. It consists of Medium and Long subsets; due to the high computational cost, we conduct experiments only on the Medium subset.

It should be noted that all HaluMem evaluation metrics are based on LLM-as-Judge. In contrast, LongMemEval provides a ground-truth list of relevant sessions for each test query and computes retrieval metrics such as recall and NDCG based on this list. Considering both metric robustness and experimental efficiency, we primarily base our method comparisons and selection across different stages on the LongMemEval dataset.

\section{Extended Experimental Results} \label{appendix_extendedexperimentalresults}
To supplement the core conclusions, this appendix offers an extended report of our experimental results from Section~\ref{exp:end2end}. Furthermore, we include ancillary analysis and exploratory comparisons to discuss the trade-offs and empirical motivations behind several minor design configurations.

\subsection{Full Evaluation Metrics}

\begin{figure*}[t!] 
  \centering  
  \begin{subfigure}{0.95\textwidth}
    \centering
    \includegraphics[width=\linewidth]{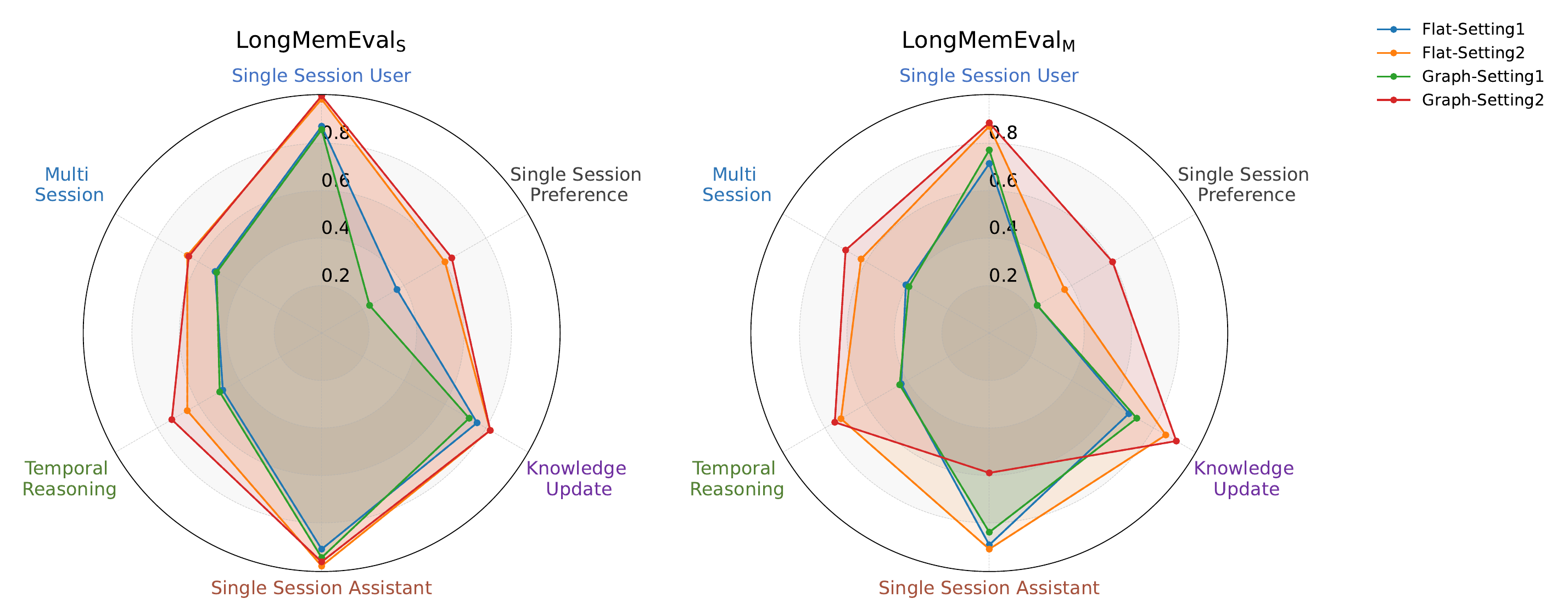}  
    \subcaption{Answer Accuracy}
  \end{subfigure}\\[6pt]  
  \begin{subfigure}{0.95\textwidth}
    \centering
    \includegraphics[width=\linewidth]{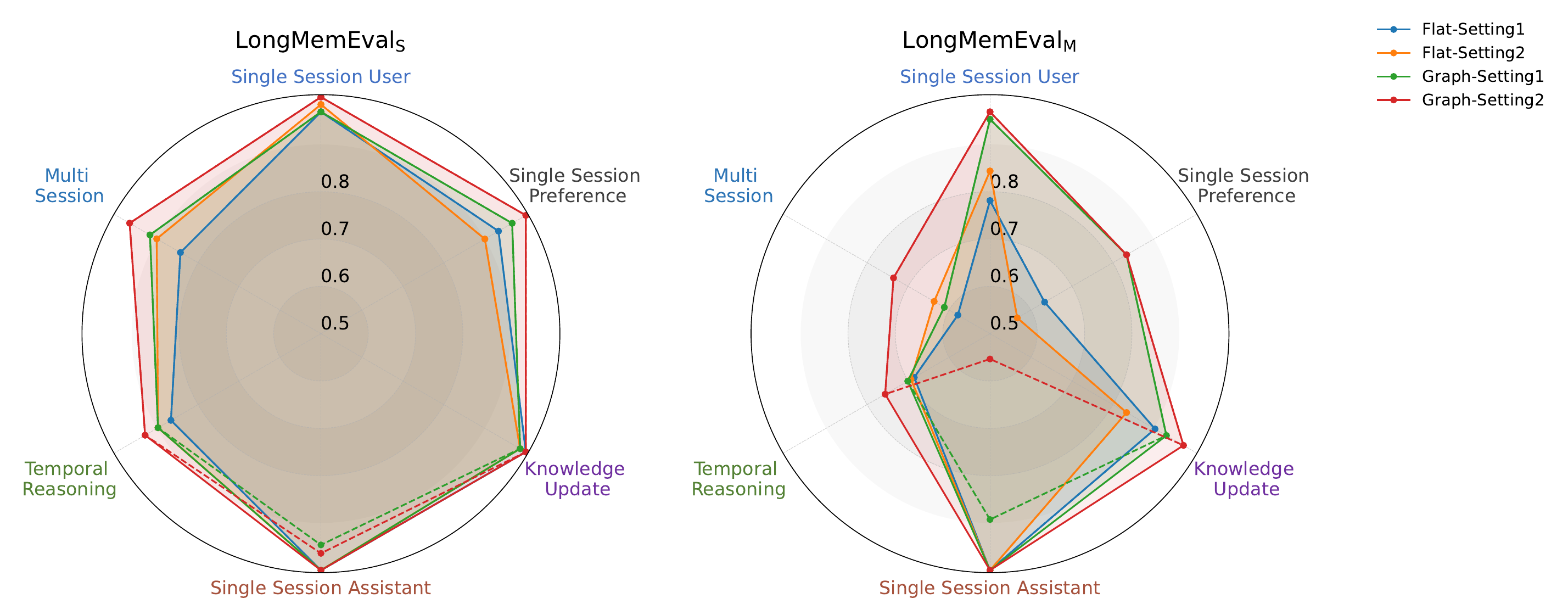}
    \subcaption{Retrieval Recall@5}
  \end{subfigure}\\[6pt]
  \begin{subfigure}{0.95\textwidth}
    \centering
    \includegraphics[width=\linewidth]{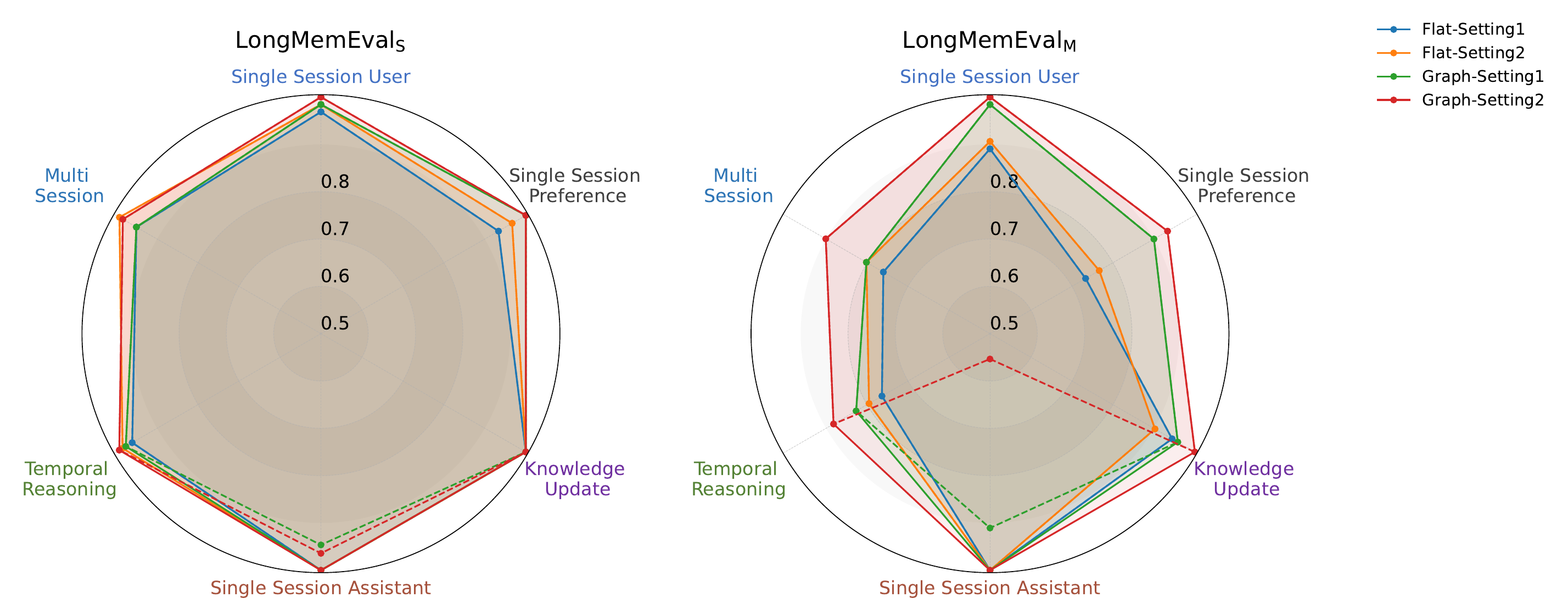}
    \subcaption{Retrieval Recall@10}
  \end{subfigure}
  \caption{\textbf{Performance of Different Configurations Across Question Types in LongMemEval.} 
The plots illustrate results under two experimental environments: 
(1) \textbf{Setting 1 (Local Deployment)}, utilizing Llama-3.1-8B for extraction/QA and Contriever for embeddings; and 
(2) \textbf{Setting 2 (API Service)}, employing GPT-4o-mini and text-embedding-3-small. 
Dashed lines indicate retrieval metrics for Single-Session Assistant questions without filtering. 
In contrast, following the standard LongMemEval protocol, the primary values reported in our main text exclude questions where the ground truth resides outside user turns.}
  \label{fig:QA_radar}
\end{figure*}

Figure~\ref{fig:QA_radar} presents the Question Answering (QA) and retrieval performance across our four primary experimental configurations on the $\mathrm{LongMemEval_{S}}$ and $\mathrm{LongMemEval_{M}}$ benchmarks. The results are visualized using radar charts to provide a comprehensive assessment of model capabilities across six distinct problem dimensions: Single Session User, Single Session Preference, Knowledge Update, Single Session Assistant, Temporal Reasoning, and Multi-Session.

The evaluation compares four experimental settings, which are consistent with the configurations detailed in Table ~\ref{tab:main_results_lme}. In this framework, "Flat" and "Graph" denote the specific indexing methodologies employed for memory organization. The term "Setting" distinguishes the scale and capacity of the Large Language Models (LLMs) used for memory construction and retrieval.
\begin{compactitem}
\item \textbf{Setting 1 (Local Deployment)}: Utilizes Llama-3.1-8B for memory extraction and QA, paired with Contriever for generating embeddings. This configuration represents a more constrained experimental environment.

\item \textbf{Setting 2 (API Service)}: Employs GPT-4o-mini for extraction and QA, alongside text-embedding-3-small for embedding tasks.
\end{compactitem}
For all experiments, GPT-4o is utilized as the backbone evaluator to ensure consistency. As Setting 1 operates under lower specifications, its performance is generally expected to be inferior to that of Setting 2.

Interestingly, Graph-setting2 exhibits an unexpected performance drop in Single Session Assistant queries — a phenomenon also noted but left unanalyzed in the Zep and Nemori~\cite{rasmussen2025zep,nan2025nemori}. We identify the  cause as a mismatch in information sourcing: while the ground truth for these questions lies in the assistant's responses, standard efficiency-driven practices—including ours—only use user messages to build the graph. Consequently, the graph itself lacks the necessary answer nodes. Our prejudge mechanism exacerbates this by filtering out messages it deems "uninformative," effectively removing the bridge between the question and the answer. Counter-intuitively, this issue becomes more severe with larger models, as their higher precision leads to stricter (and thus more destructive) filtering. This presents a clear trade-off: the prejudge mechanism is essential for reducing overhead and boosting overall system recall, but it costs us performance in this specific sub-type. To maintain global efficiency, we ultimately retained this approach.

Table~\ref{tab:halumem_fulltable} presents the complete experimental results on the HaluMem dataset, serving as a supplement to Table~\ref{tab:main_results_halumem} in the main text. 
We observe that the strong baselines we propose consistently outperform the results reported in Table~\ref{tab:halumem_fulltable}, particularly on the Memory Recall metric. Since the HaluMem paper does not report the detailed experimental settings of each method (e.g., the choice of memory extraction models and embedding models), it is difficult to attribute the observed performance differences.

\begin{table*}[t]
\centering
\small
\begin{tabular}{l l l l c c c c c c}
\toprule
\multirow{2}{*}{Model Setting} &
\multirow{2}{*}{Key Org} &
\multirow{2}{*}{Operation} &
\multirow{2}{*}{Edge} &
\multicolumn{3}{c}{Extraction} &
\multicolumn{3}{c}{Answer (V=Key)} \\
\cmidrule(lr){5-7} \cmidrule(lr){8-10}
& & & &
Mem-R & Mem-P & MemUpdate-C &
QA-C & QA-H & QA-O \\
\midrule
Local Deploy & \texttt{s,f,k} & A & N
  & 0.7332 & 0.9360 & - & 0.5492 & 0.1263 & 0.3245 \\
API service & \texttt{s,f,k} & A & N
  & 0.7686 & 0.8926 & - & 0.6637 & 0.1526 & 0.1837 \\

Local Deploy & \texttt{session,S,F,K} & A/U/N & N
  & 0.7121 & 0.9811 & 0.1095 & 0.4800 & 0.2157 & 0.3043 \\
Local Deploy & \texttt{[session,S,F,K]} & A/U/N & N
  & 0.7124 & 0.9808 & 0.1537 & 0.5535 & 0.1552 & 0.2913 \\
Local Deploy & \texttt{[session,S,F,K]} & A/U/N & Y
  & 0.7124 & 0.9808 & 0.1537 & 0.5524 & 0.1635 & 0.2841 \\

\bottomrule
\end{tabular}
\caption{External flat-based results on HaluMem-medium. Key=\texttt{s,f,k} refers to Separate organization; Key=\texttt{session,S,F,K} refers to Merge-by-type organization; Key=\texttt{[session,S,F,K]} refers to Merge-by-all organization. A=Add, U=Update, N=Noop.}
\label{tab:all_results_halumem}
\end{table*}

\begin{table*}[htbp]
\centering
\tiny
\setlength{\tabcolsep}{5pt} 
\begin{tabular}{ll ccc ccc ccc ccc}
\toprule
\multirow{2}{*}{\textbf{Dataset}} & \multirow{2}{*}{\textbf{System}} & \multicolumn{2}{c}{Memory Integrity} & \multicolumn{3}{c}{Memory Accuracy} & \multicolumn{3}{c}{Memory Updating} & \multicolumn{3}{c}{Question Answering} \\ 
\cmidrule(lr){3-4} \cmidrule(lr){5-7} \cmidrule(lr){8-10} \cmidrule(lr){11-13}
& & R & Weighted R & Target P & Acc. & FMR & C & H & O & C & H & O \\ \midrule

\multirow{5}{*}{Medium} 
& Mem0 & 42.91\% & 65.03\% & 86.26\%(10556) & 60.86\%(16291) & 56.80\% & 25.50\% & 0.45\% & 74.02\% & 53.02\% & \textbf{19.17\%} & 27.81\% \\
& Mem0-Graph & \textbf{43.28\%} & \textbf{65.52\%} & 87.20\%(10567) & \textbf{61.86\%(16230)} & 55.70\% & 24.50\% & 0.26\% & 75.24\% & 54.66\% & 19.28\% & 26.06\% \\
& Memobase & 14.55\% & 25.88\% & \textbf{92.24\%(5443)} & 32.29\%(17081) & \textbf{80.78\%} & 5.20\% & 0.55\% & 94.25\% & 35.33\% & 29.97\% & 34.71\% \\
& Supermemory & 41.53\% & 64.76\% & 90.32\%(14134) & 60.83\%(22551) & 51.77\% & 16.37\% & 1.15\% & 82.47\% & 54.07\% & 22.24\% & 23.69\% \\
& Zep & - & - & - & - & - & \textbf{47.28\%} & 0.42\% & \textbf{52.31\%} & \textbf{55.47\%} & 21.92\% & \textbf{22.62\%} \\ \midrule

\multirow{5}{*}{Long} 
& Mem0 & 3.23\% & 11.89\% & 88.01\%(1134) & \textbf{46.01\%(2433)} & 87.65\% & 1.45\% & 0.03\% & 98.51\% & 28.11\% & \textbf{17.29\%} & 54.60\% \\
& Mem0-Graph & 2.24\% & 10.76\% & 87.32\%(785) & 41.26\%(1866) & \textbf{88.36\%} & 1.47\% & 0.04\% & 98.40\% & 32.44\% & 21.82\% & 45.74\% \\
& Memobase & 6.18\% & 14.68\% & \textbf{88.56\%(3077)} & 25.61\%(11795) & 85.39\% & 4.10\% & 0.36\% & 95.38\% & 33.60\% & 29.46\% & 36.96\% \\
& Supermemory & \textbf{53.02\%} & \textbf{70.73\%} & 85.82\%(24483) & 29.71\%(77134) & 36.86\% & 17.01\% & 0.58\% & 82.42\% & \textbf{53.77\%} & 22.21\% & \textbf{24.02\%} \\
& Zep & - & - & - & - & - & \textbf{37.35\%} & 0.48\% & \textbf{62.14\%} & 50.19\% & 22.51\% & 27.30\% \\ \bottomrule
\end{tabular}

\vspace{0.2cm}
\raggedright
\footnotesize
\caption{Original evaluation results reported in HaluMem paper.}
\label{tab:halumem_fulltable}
\end{table*}

\begin{table*}[t]
\centering
\small
\begin{tabular}{l l c c c c}
\toprule
\multirow{2}{*}{Model Setting} &
\multirow{2}{*}{Key Org} &
\multicolumn{2}{c}{Retrieval} &
\multicolumn{2}{c}{Answer Accuracy} \\
\cmidrule(lr){3-4} \cmidrule(lr){5-6}
& &
R@5 & R@10 &
V=session & V=Key \\
\midrule

\rowcolor{gray!15}
\multicolumn{6}{c}{\textbf{LongMemEval-S}} \\
\midrule
Local Deploy & \texttt{session,[S,F,K]}
  & 0.9379 & 0.9833 & 0.638 & 0.594 \\
API service & \texttt{session,[S,F,K]}
  & 0.9498 & 0.9904 & 0.756 & 0.774 \\

\midrule
\rowcolor{gray!15}
\multicolumn{6}{c}{\textbf{LongMemEval-M}} \\
\midrule
Local Deploy & \texttt{session,[S,F,K]}
  & 0.7685 & 0.8592 & 0.562 & 0.496 \\
API service & \texttt{session,[S,F,K]}
  &  0.7661 & 0.8592 & 0.660 & 0.660 \\

\bottomrule
\end{tabular}
\caption{External flat-based results on LongMemEval. Key=\texttt{session,[S,F,K]} refers to refined Merge-by-session organization.}
\label{tab:all_results_lme}
\end{table*}


\subsection{Cost-Efficiency Analysis}

\begin{table*}[htbp]
\centering
\footnotesize
\setlength{\tabcolsep}{4pt}
\renewcommand{\arraystretch}{1.2}
\begin{tabular}{llllllll}
\toprule
Dataset &
U Session \# &
U Session tok. &
Method &
Memory \# &
Memory tok. &
Extr. (min) &
Retr. (ms/q) \\
\midrule

\multirow[c]{2}{*}{LME-s} &
\multirow[c]{2}{*}{19.2k} &
\multirow[c]{2}{*}{41.6m} &
flat &
\makecell[l]{kw: 375.9k \\ fact: 111.8k \\ sum: 19.2k} &
\makecell[l]{kw: 1.3m \\ fact: 1.9m \\ sum: 2.0m} &
156 &
45 \\
\cmidrule(lr){4-8}
& & &
graph &
\makecell[l]{node: 156.1k \\ edge: 145.9k} &
desc: 3.6m &
1181 &
44 \\
\midrule

\multirow[c]{2}{*}{LME-m} &
\multirow[c]{2}{*}{51.6k} &
\multirow[c]{2}{*}{107.6m} &
flat &
\makecell[l]{kw: 1,042k \\ fact: 295.1k \\ sum: 51.6k} &
\makecell[l]{kw: 3.5m \\ fact: 5.0m \\ sum: 5.5m} &
452 &
240 \\
\cmidrule(lr){4-8}
& & &
graph &
\makecell[l]{node: 1,218k \\ edge: 1,266k} &
desc: 27.6m &
1781 &
574 \\
\midrule

\multirow[c]{2}{*}{HaluMem-m} &
\multirow[c]{2}{*}{1387} &
\multirow[c]{2}{*}{3.2m} &
flat &
\makecell[l]{kw: 32.7k \\ fact: 19.3k \\ sum: 1387} &
\makecell[l]{kw: 101k \\ fact: 405k \\ sum: 146k} &
308 &
15 \\
\cmidrule(lr){4-8}
& & &
graph &
\makecell[l]{node: 14.8k \\ edge: 14.6k} &
desc: 438k &
673 &
93 \\
\bottomrule
\end{tabular}
\caption{Statistics and efficiency of flat and graph memory construction}
\label{tab:memory_stats}
\end{table*}

Table~\ref{tab:memory_stats} summarizes the computational cost of flat and graph-based dialog memory systems under practical implementation settings.
Statistics are reported over \emph{unique} dialog sessions and tokens, as memory extraction is cached across repeated content.
All experiments use \texttt{gpt-4o-mini} for memory extraction, \texttt{text-embedding-3-small} for embedding, and multi-threaded execution with \texttt{max\_workers=16}.

\paragraph{Retrieval latency.}
Retrieval latency directly affects user experience and is therefore the most critical efficiency metric.
Graph-based retrieval is consistently slower than flat retrieval, but remains within a practically acceptable range.
On LongMemEval-S, both methods exhibit comparable latency (45 vs.\ 44 ms per query).
On LongMemEval-M, graph retrieval incurs 574 ms per query compared to 240 ms for flat memory.
Even at this scale (over 50k unique sessions), retrieval latency remains on the order of a few hundred milliseconds, which is typically acceptable for interactive dialog systems.

\paragraph{Memory extraction cost.}
Memory extraction is usually not on the critical path of user interaction and is often performed asynchronously after a session ends.
Graph-based memory construction is more expensive than flat memory due to additional processing such as entity extraction and description aggregation.
However, when normalized by the number of unique sessions, the cost remains manageable: on LongMemEval-M, graph extraction takes about 2.1 seconds per session, compared to 0.5 seconds for flat memory.

\paragraph{Memory storage.}
Both flat and graph-based memory achieve substantial compression relative to raw dialog content.
Although graph memory stores more tokens than flat memory, the difference is modest compared to the original dialog size.
For example, on LongMemEval-M, over 100M tokens of dialog are compressed into fewer than 30M tokens of stored memory in both settings.

\paragraph{Summary.}
Flat memory offers lower construction and retrieval cost, while graph-based memory introduces additional overhead that increases with scale.
Nevertheless, this overhead remains manageable under realistic usage patterns, suggesting that efficiency alone should not preclude the adoption of graph-based designs when they provide modeling benefits.

\subsection{Supplementary Design Experiments}
In this subsection, we delve into the empirical rationale behind several design choices. By comparing alternative configurations, we justify the specific settings adopted in our final framework.
\subsubsection{Key Selection on HaluMem}
Table~\ref{tab:key_halumem_ablation} shows the results of key design experiments conducted on HaluMem. To perform the retrieval recall analysis on HaluMem, we adopted the evaluation protocol of LongMemEval. Since HaluMem does not natively provide the source session for each question's evidence, we constructed the ground truth by matching the evidence with the memory points across all sessions.

As shown in Table~\ref{tab:key_halumem_ablation}, the results on HaluMem remain consistent with our findings in Table~\ref{tab:key_ablation}. The introduction of refined keys via the separate strategy (session, S, F, K) yields superior results in top-5 retrieval recall. Conversely, the merge strategy often leads to performance degradation, with [session, S, F, K] even underperforming the session-only baseline.

When raw session content is unavailable, the separate strategy (S, F, K) significantly outperforms the merge strategy ([S, F, K]). This confirms that maintaining independent keys is more robust, especially under constrained retrieval settings.

\begin{table}[t]
\centering
\small
\setlength{\tabcolsep}{4pt}
\begin{tabular}{lcc}
\toprule
Key Design & R@5 & R@10  \\
\midrule
\texttt{session} & 0.4739 & 0.6002 \\
\texttt{session,S,F,K} & 0.4860& 0.5988  \\
\texttt{[session,S,F,K]} & 0.4716& 0.5994  \\
\texttt{session,[S,F,K]} & 0.4742& 0.5924  \\
\midrule
\texttt{S,F,K} & 0.4609& 0.5795  \\
\texttt{[S,F,K]} & 0.4211& 0.5160 \\
\bottomrule
\end{tabular}
\caption{Retrieval performance on HaluMem under different keys. S=summary, F=factual statement, K=keyword. \texttt{[a,b]} means their representations are merged during retrieval. R@n means Recall@n.}
\label{tab:key_halumem_ablation}
\end{table}

\subsubsection{Ablation Study on the Add Operation}
In Section~\ref{exp:operation}, we primarily investigated two memory update strategies: \textit{add} and \textit{add/update/noop}. To further evaluate the specific contribution of the \texttt{add} operation to the overall system performance, we conducted an ablation study as presented in Table~\ref{tab:op_add_ablation}.

In this experiment, we compare two settings under the Separate(S, F, K) indexing strategy: 
\begin{compactitem} 
    \item \textbf{w/ add}: A comprehensive update strategy including \texttt{add}, \texttt{update}, and \texttt{noop}. 
    \item \textbf{w/o add}: A restricted strategy that only allows \texttt{update} and \texttt{noop}. 
\end{compactitem}

The results indicate that while memory extraction metrics remain relatively stable, the inclusion of the \texttt{add} operation significantly enhances memory update accuracy and final end-to-end QA performance. This underscores the necessity of the \texttt{add} operation for maintaining an up-to-date and comprehensive memory index. Consequently, the ``w/ add'' configuration demonstrates superior overall performance; thus, we adopt this setting for all update operations discussed in the main text.

\begin{table}[t]
\centering
\small
\setlength{\tabcolsep}{4pt}
\begin{tabular}{lcccc}
\toprule
Update Op & Mem-R & Mem-P & MemUpdate-C & QA-C \\
\midrule
\texttt{w add} & 0.7121& 0.9811& 0.1095& 0.4800\\
\texttt{w/o add} & 0.7185& 0.9814& 0.0332& 0.3754 \\
\bottomrule
\end{tabular}
\caption{Ablation results of the \texttt{add} operation on the HaluMem-medium dataset. "w/ add" denotes the full operation set (\textit{add/update/noop}), while "w/o add" refers to the restricted set (\textit{update/noop}). \textit{Mem-R}, \textit{Mem-P}, \textit{MemUpdate-C}, and \textit{QA-C} represent Memory Extraction Recall, Target Memory Extraction Precision, Correct Ratio of memory update operations, and end-to-end QA Accuracy, respectively.}
\label{tab:op_add_ablation}
\end{table}

\subsubsection{Ablation on Prejudge Mechanism}
The Prejudge Mechanism is a strategic filtering step integrated into our memory construction pipeline. For each incoming text chunk, the system employs an LLM to evaluate its informativity before performing fine-grained extraction. Chunks deemed irrelevant or redundant are immediately discarded, while only those identified as valuable proceed to the subsequent information extraction and graph update stages.

As shown in Table~\ref{tab:index_prejudge}, we compare the performance with the Prejudge Mechanism enabled (\texttt{w/}) and disabled (\texttt{w/o}). The results demonstrate that incorporating Prejudge does not compromise retrieval or ranking performance; in fact, it yields slight improvements across all metrics. By filtering out noise at the source, the mechanism effectively maintains a high signal-to-noise ratio in the memory index while substantially reducing unnecessary computational overhead for the subsequent update and storage modules. Consequently, the Prejudge Mechanism is employed in all experimental configurations throughout this paper unless otherwise noted.

\begin{table}[t]
\centering
\small
\setlength{\tabcolsep}{4pt}
\begin{tabular}{lcccc}
\toprule
Configuration & R@5 & R@10 & N@5 & N@10 \\
\midrule
\texttt{w/o Prejudge} & 0.9284 & 0.9689 & 0.9308 & 0.9386 \\
\texttt{w/ Prejudge} & \textbf{0.9355} & \textbf{0.9761} & \textbf{0.9392} & \textbf{0.9476} \\
\bottomrule
\end{tabular}
\caption{Ablation results of the Prejudge Mechanism. R@$k$ and N@$k$ denote Recall@$k$ and nDCG@$k$, respectively. The results indicate that the Prejudge Mechanism optimizes computational efficiency without compromising (and even slightly improving) retrieval performance.}
\label{tab:index_prejudge}
\end{table}

\section{Comparison with Related Works}\label{appendix_relatedsystem}
\subsection{Breakdown of Related Systems} 
Table~\ref{tab:memory_comparison_fulltable} decomposes and compares several representative dialog memory systems under a unified framework, the abbreviations and symbols used in the column headers and entries are explained below:   

\begin{compactitem}
    \item \textbf{Key, Value:} Summ: Summary, Kw: Keyword, Ins: Insight, Comm: Community.
    \item \textbf{Query:} Q: Query, Q+T: Query + Time, Rew Q: Rewrited Query
     \item \textbf{Index Structure:} F: Flat, G: Graph, HG: Hybrid Graph, FG: Flat+Graph.
    \item \textbf{Index Operations:} Add: Add, Upd: Update, Align: Alignment, Comm Upd: Community Update,  N\texttt{/}E Align: Node\texttt{/}Edge Alignment.
    \item \textbf{Retrieval:} $\rightarrow$: process flow, 1-Hop: one-hop neighbor retrieval, Rerank: re-ranking, PPR: Personalized PageRank, Kw: Keyword.
    \item \textbf{Answering:} CoN: Chain-of-note Answering, Direct: Direct Answering.
\end{compactitem}

\textbf{Interpretation Note:} The table~\ref{tab:memory_comparison_fulltable} decomposes and compares the design choices across systems under our proposed unified framework. The \textit{Key}, \textit{Value}, and \textit{Index} columns reveal how memory is organized. The \textit{Retrieval} column depicts the search path complexity, and the \textit{Query} \textit{Answering} column distinguishes response generation methods.

\begin{table*}[htbp]
\centering
\setlength{\tabcolsep}{4pt}
\tiny
\begin{tabular}{lllllll}
\toprule
 & \textbf{Key} & \textbf{Value} & \textbf{Query} & \textbf{Index (Struct; Op)} & \textbf{Retrieval} & \textbf{Answering} \\
\midrule
\textbf{LongMemEval} & Session + Fact & Session & Q + T & F; Add & Q $\rightarrow$ K $\rightarrow$ V & CoN \\

\textbf{RMM} & Topic summ & Session + Key& Q& F; Add, Upd & Q $\rightarrow$ K $\rightarrow$ V $\rightarrow$ Rerank & Direct \\

\textbf{A-Mem} & Session + Kw, Tag, Summ& Key & Q & G; Add, Upd & Q $\rightarrow$ K $\rightarrow$ V & Direct \\

\textbf{Mem0-G} & Entity name, Triple & Triple & Q& G; Add, Align & Q $\rightarrow$ K $\rightarrow$ 1-Hop $\rightarrow$ V& Direct \\

\textbf{Zep} & Entity summ, Triple, Comm & Key & Q & HG; Add, Align, Comm Upd & Q $\rightarrow$ K $\rightarrow$ 1-Hop $\rightarrow$ V $\rightarrow$ Rerank& Direct \\

\textbf{SGMem} & Sentence, Summ, Ins, Fact& Key & Q & FG; Add& Q $\rightarrow$ K $\rightarrow$ 1-Hop $\rightarrow$ V& Direct \\

\textbf{Nemori} & Episode+Semantic Memory& Key & Q & F; Add, Upd & Q $\rightarrow$ K $\rightarrow$ V& Direct \\

\textbf{CAM} & Chunk, Comm& Key& Q & HG; Add, Comm Upd& Q $\rightarrow$ K $\rightarrow$ 1-Hop $\rightarrow$ V& Direct \\

\textbf{LightRAG} & Entity name, Triple& Chunk + Key& Q $\rightarrow$ Kw& G; Add, Align,& Q $\rightarrow$ Kw $\rightarrow$ Entity+Triple$\rightarrow$1-Hop $\rightarrow$ V& Direct \\

\textbf{EcphoryRAG} & Entity, Chunk& Key & Q & G; Add& Q $\rightarrow$ Entity+Chunk $\rightarrow$ 1-Hop $\rightarrow$ V& Direct \\

\textbf{HippoRAG2} & Entity, Triple, Chunk& Chunk& Rew Q & G; Add, N Align& Q $\rightarrow$ Triple  $\rightarrow$ Node $\rightarrow$PPR Rerank$\rightarrow$ V& Direct \\

\textbf{AriGraph} & Triple& Chunk& Q & G; Add, E Align& Q $\rightarrow$ Triple $\rightarrow$ Node $\rightarrow$ 1-Hop $\rightarrow$Triple$\rightarrow$ V& Direct \\

\bottomrule
\end{tabular}
\caption{Decomposition of some representative dialog memory systems under the unified framework.}
\label{tab:memory_comparison_fulltable}
\end{table*}

\subsection{Comparison of System Settings}

Table~\ref{tab:memory_misalignment} summarizes the experimental settings and reported results of several representative dialog memory and RAG frameworks. Our analysis reveals two primary observations:

 \noindent \textbf{1. Misalignment in Evaluation Environments.} There is a significant lack of uniformity in the foundational components used across different studies. As shown in the table, existing methods vary widely in their choice of information extraction models (ranging from Llama-3.1 to Gemini-1.5-Flash), embedding models (e.g., Stella V5, BGE-m3, text-embedding-3), and top-$k$ retrieval constraints. Furthermore, the Value Type stored in memory—varying from raw Sessions and Facts to specific Graph Nodes or Episodic/Semantic units—further complicates direct performance comparisons. Such discrepancies in "system-level" configurations make it difficult to isolate the true effectiveness of any specific memory architecture or retrieval paradigm.
 
 \noindent \textbf{2. Performance Superiority via Systemic Refinement.} Despite these inconsistent benchmarks, our framework achieves state-of-the-art performance across all metrics. Notably, beyond standard retrieval recall (0.969 on LME-S), we emphasize our end-to-end QA accuracy, which reaches 0.892 and 0.754 using gpt-4o-mini on the S and M subsets, respectively. These results significantly surpass contemporary baselines, demonstrating that our gains are not merely a byproduct of backbone selection but a result of our systematic deconstruction of the dialog memory pipeline. By meticulously optimizing each modular component—from refined key design to update logic—our engineering-focused implementation proves that a unified and well-calibrated framework can overcome the inherent complexities of long-term memory management.

In summary, the substantial performance margins reported in Table~\ref{tab:main_results_lme} underscore the effectiveness of our design choices, providing a robust and reproducible benchmark for future research in long-term dialog memory systems.

\begin{table*}[t]
\centering
\setlength{\tabcolsep}{3pt}
\tiny
\begin{tabularx}{\textwidth}{l p{50pt} X p{45pt} c l c cc cc}
\toprule
 & & & & & & & \multicolumn{2}{c}{\textbf{LongMemEval-S}} & \multicolumn{2}{c}{\textbf{LongMemEval-M}} \\
\cmidrule(lr){8-9} \cmidrule(lr){10-11}
\textbf{Model} & \textbf{Extraction} & \textbf{Embedding} & \textbf{QA Model} & \textbf{top-k} & \textbf{Value Type} & \textbf{Eval.} & \textbf{Recall} & \textbf{QA Acc.} & \textbf{Recall} & \textbf{QA Acc.} \\
\midrule
LongMemEval & Llama-3.1-8B & Stella V5 & gpt-4o & 5/10 & Session & gpt-4o & $-$ & $-$ & 0.732/0.862 & 0.714/0.700 \\
RMM & Gem-1.5-Flash & GTE & Gem-1.5-Flash & 5/10 & Session & Gem-1.5-Pro & $-$ & $-$ & 0.698/0.744 & 0.704/0.738 \\
A-mem & $-$ & $-$ & $-$ & $-$ & $-$ & $-$ & 0.626/0.652* & $-$ & $-$ & $-$ \\
Mem0-g & $-$ & $-$ & $-$ & $-$ & $-$ & $-$ & 0.536/0.395* & $-$ & $-$ & $-$ \\
Zep & gpt-4o-mini & BGE-m3 & 4o-mini/4o & 20 & Fact+Node & gpt-4o & 0.638/0.712 & $-$ & $-$ & $-$ \\
SG-mem & Qwen2.5-32B & S-BERT & Qwen2.5-32B & 5/10 & Session & Qwen2.5-32B & $-$ & $-$ & $-$ & 0.700/0.730 \\
Nemori & gpt-4o-mini & text-emb-3 & gpt-4o-mini & 10+20 & Epi+Sem & gpt-4o-mini & $-$ & 0.744/0.794 & $-$ & $-$ \\
LightMem & $-$ & $-$ & 4o-mini/Qwen3 & $-$ & $-$ & $-$ & $-$ & 0.686/0.702 & $-$ & $-$ \\
\midrule
\textbf{Ours} & gpt-4o-mini & text-emb-3 & gpt-4o-mini & 5 & Session & gpt-4o & \textbf{0.969/0.935} & \textbf{0.892/0.620} & \textbf{0.828/0.766} & \textbf{0.754/0.548} \\
\bottomrule
\end{tabularx}
\vspace{2pt}
\begin{flushleft}
\tiny \textbf{Notes}: \textbf{4o-mini/4o}: gpt-4o-mini/gpt-4o; \textbf{Qwen3}: Qwen3-30B-A3B-Instruct; \textbf{Gem-1.5-Flash/Pro}: Gemini-1.5-Flash/Pro; \textbf{text-emb-3}: text-embedding-3-small; \textbf{BGE-m3}: BAAI General Embedding; \textbf{S-BERT}: Sentence-BERT(all-MiniLM-L6-v2); \textbf{GTE}: General Text Embeddings; \textbf{Epi+Sem}: Episodic + Semantic Memory.Results marked with* are those reported in the LingtMem~\cite{fang2025lightmem} paper.
\end{flushleft}
\caption{Comparison of various dialog memory systems and RAG frameworks. results are collected from original papers/reports.}
\label{tab:memory_misalignment}
\end{table*}

\section{Case Study} \label{appendix_casestudy}
This section presents several case studies.
\newtcolorbox{widecasebox}[1]{
    colback=white, 
    colframe=black!70, 
    fonttitle=\bfseries,
    title=#1,
    arc=0mm,
    boxrule=0.8pt,
    width=\textwidth, 
    left=3mm, right=3mm, top=2mm, bottom=2mm
}
\subsection{Case study: Graph Index Failure Cases}
As illustrated in Table \ref{tab:main_results_lme}, while the graph-based index achieves superior retrieval performance, this advantage does not consistently translate into downstream QA accuracy. Specifically, in the $V=\text{Session}$ setting, the graph method performs on par with the flat baseline, whereas in the $V=\text{Key}$ setting, it even underperforms. To investigate the underlying causes of this discrepancy, we conduct a qualitative analysis of representative cases where the graph strategy fails despite successful retrieval.

Figure \ref{fig:appendix_badcase1} presents a typical failure in the $V=\text{Session}$ configuration. In this instance, the graph-based index retrieves and presents more time-related metadata within the prompt context. Analysis shows that for a smaller-scale model like Llama-3.1-8B, these additional temporal details act as distracting information. The model is indeed capable of recognizing the target fact (e.g., the specific internet speed of 500 Mbps) from the text; however, its reasoning process is misled by the salient yet irrelevant temporal noise, causing it to fail in selecting and outputting the correct answer. In contrast, the flat index strategy adopts a more concise organization of index, presenting context in a direct, narrative-driven manner. This approach not only enhances the expression of semantic information but also significantly reduces interference from irrelevant timestamp details, thereby retrieving the most semantic relevant information. This enables the model to maintain focus on the semantic core and successfully output the correct answer.

Figure \ref{fig:appendix_badcase2} reveals a failure mechanism in the $V=\text{Key}$ setting. Here, the graph-based index retrieves 20 atomic entities (e.g., "TRIPLE THE PRICE", "FLEA MARKET"). Although relevant, this highly fragmented representation strips away essential descriptive context and semantic associations between entities. Consequently, the model fails to perform entity alignment—for example, it cannot confidently link the "painting of a sunset" in the query to the "flea market find" in the retrieved facts.In contrast, the flat strategy generates semantically cohesive "memory notes" by integrating key facts with multi-faceted summaries. Thus, the context provided by 20 memory notes carries significantly more relational information than 20 isolated entities. This structured representation preserves sufficient narrative anchors and semantic cohesion, enabling the model to bridge the semantic gap between the query and the evidence, even in the presence of price-related distractors.

Through these case studies, we attribute the decline in QA performance to two key factors. First, for capacity-limited models, organizing the retrieved context in a semantically coherent and structurally clear manner is crucial for downstream QA performance. Second, while the atomic entities and their concise descriptions used by the graph-based index contribute to its interpretability, in practice, this very design often results in informationally sparse representations, creating a performance bottleneck that must be carefully considered.

\begin{figure*}[t]
\centering
\begin{tcolorbox}[
    colback=white, 
    colframe=black!75, 
    fonttitle=\bfseries,
    title={Case Study: Graph fail case(ID: ad7109d1)},
    arc=0mm,
    boxrule=0.8pt,
    width=1.0\textwidth,
    left=2mm, right=2mm, top=2mm, bottom=2mm
]
    \small
    \textbf{User Query:} \textit{``What speed is my new internet plan?''} \\
    \textbf{Answer:} 500 Mbps\\
    \textbf{Ground Truth Session:}
    "user":I'm having some issues with my laptop's battery life. Can you give me some tips on how to extend it? "assistant":...
    "user":I did notice that my internet speed has been really good lately, especially when I'm streaming movies on Netflix. I upgraded to \textbf{500 Mbps} about three weeks ago, and it's made a huge difference. Anyway, back to my battery issue... do you think it's possible that my laptop's battery is just old and needs to be replaced?
    ...

    \vspace{2mm}
\begin{tabularx}{\textwidth}{X | X}
        \rowcolor{gray!10}
        \centering \textbf{Graph  Index} & \centering \textbf{Flat index} \tabularnewline
        \midrule
        \textbf{Retrieved Sessions:} & \textbf{Retrieved Sessions:} \\
        $\bullet$ \textbf{Session 1 (Target):} \textit{...I upgraded to 500 Mbps about three weeks ago...} & $\bullet$ \textbf{Session 2 (Target):} \textit{...I upgraded to 500 Mbps about three weeks ago...} \\
        $\bullet$ \textbf{Session 2-3:} [Irrelevant: Maintenance schedules and shopping logs with heavy timestamps.] & $\bullet$ \textbf{Session 1, 3-5:} [Irrelevant: Miscellaneous topics including yoga, meal prep, and shuttle services.] \\
        $\bullet$ \textbf{Session 4-5:} [Irrelevant: Distant historical logs regarding travel and sports.] & \\
        \addlinespace[1mm]
        \hdashline
        \addlinespace[1mm]
        {\color{red!70!black} \textbf{Model Response (Hallucination):}} & {\color{green!40!black} \textbf{Model Response (Success):}} \\
        \textit{``To answer your question...The history mentions you upgraded to 500 Mbps, \textbf{but it doesn't mention your current plan}...Can you please provide more context or information...''} & \textit{``To answer your question...In Session 2, the user mentions that they upgraded to 500 Mbps about three weeks ago,...\textbf{The user explicitly mentions that they upgraded to 500 Mbps}, which is the speed of their new internet plan.Answer: Your new internet plan is 500 Mbps.''} \\
    \end{tabularx}
\end{tcolorbox}
\vspace{-8pt}
\caption{\textbf{A representative case where the flat index succeeds while the graph index fails.} In this case, because the graph-based index retrieves more time-related information, such content introduces distracting information into the prompt context, ultimately causing the less capable Llama-3.1-8B model to fail in providing the correct answer. In contrast, the higher information density of the flat index provides a richer semantic context, allowing the model to focus on the core information without being distracted by fragmented metadata."}
\label{fig:appendix_badcase1}
\end{figure*}

\begin{figure*}[t]
\centering
\begin{tcolorbox}[
    colback=white, 
    colframe=black!75, 
    fonttitle=\bfseries,
    title={Case Study: Graph fail case(ID: b86304ba)},
    arc=0mm,
    boxrule=0.8pt,
    width=1.0\textwidth,
    left=2mm, right=2mm, top=2mm, bottom=2mm
]
    \small
    \textbf{User Query:} \textit{``How much is the painting of a sunset worth in terms of the amount I paid for it?''} \\
    \textbf{Answer:} The painting is worth triple what I paid for it.\\
    \textbf{Ground Truth Session:}
    "user":I'm thinking of taking an art history course to learn more about the context and significance of different art movements. Can you recommend any good online resources or textbooks for beginners? "assistant":...
    "user":That's really helpful. I was thinking about my flea market find, and I realized that it's actually worth triple what I paid for it, which is amazing!
    ...

    \vspace{2mm}
\begin{tabularx}{\textwidth}{X | X}
        \rowcolor{gray!10}
        \centering \textbf{Graph Index} & \centering \textbf{Flat Index} \tabularnewline
        \midrule
        \textbf{Retrieved Entities (total 20 entities):} & \textbf{Retrieved noes (total 20 memory notes):} \\
        $\bullet$ \textbf{Entity 3:} ``triple the price'': The flea market find is worth triple what the user paid for it. & $\bullet$ \textbf{Note 1:} Keywords:Organizing rare items;... Facts: The user is trying to organize... ;The user has been lucky with their flea market find and thinks it's \textbf{worth triple what they paid for it}. Summary: The user is interested in taking an art history... \\\\
        $\bullet$ \textbf{Entity 8, 11:} ``flea market'': Location where the user acquired their art piece.''flea market find'': A piece of art acquired by the user from a flea market.& $\bullet$ \textbf{Note 2:} Keywords: Organizing rare items;... Facts: The user is trying to organize and research...Summary: A user is seeking help from an AI assistant to ...\\
        $\bullet$ \textbf{Entity 4, 13, 20:} [Price Distractors]: ``\$250'' for a stand mixer;''\$20''for a vinyl record;``\$5.25'' for a bundle. & $\bullet$ \textbf{Note 3:} Keywords: Here are the keyphrases from our conversation:...Facts: The user has a vintage watch collection;...Summary: The user is a collector with a diverse collection of items...\\
        $\bullet$ \textbf{Other 14 entities:} [Irrelevant Entities]:materials used, artist reputation, living room, textbooks... & $\bullet$ \textbf{Other 17 Notes:} [Irrelevant]:note 4:Seoul trip planning note5:Stand mixer ...[others ommited] \\
        \addlinespace[1mm]
        \hdashline
        \addlinespace[1mm]
        {\color{red!70!black} \textbf{Model Response (Failure):}} & {\color{green!40!black} \textbf{Model Response (Success):}} \\
        \textit{``To answer the question...3: TRIPLE THE PRICE: The flea market find is worth triple what the user paid for it....However, there is no specific information about the painting of a sunset....To clarify, I would like to ask:...''} & \textit{``To answer the question... The user has been lucky with their flea market find and thinks it’s worth triple what they paid for it.... Answer: It is worth triple what you paid.''} \\
    \end{tabularx}
\end{tcolorbox}
\vspace{-8pt}
\caption{\textbf{A representative case study comparing retrieval strategies in Value=Key setting.} While the graph-based index retrieves 20 atomic entities, their fragmented nature strips away essential context, leading the model to fail in entity alignment (e.g., linking the "flea market find" to the "painting"). In contrast, the flat index succeeds by providing 20 memory notes that integrate keywords, facts, and summaries. This structured representation preserves sufficient semantic cohesion, allowing the model to focus on the semantic core and reason correctly despite the presence of irrelevant distractors.}
\label{fig:appendix_badcase2}
\end{figure*}

\subsection{Case study: Graph Index Success Cases}
To provide a comprehensive view, we further examine representative success cases of the graph index strategy in Figure~\ref{fig:appendix_goodcase1} and Figure~\ref{fig:appendix_goodcase2}, both evaluated using the Llama3.1-8B model on the LongMemEval-S and LongMemEval-M benchmarks, respectively.

The phenomenon in Figure~\ref{fig:appendix_goodcase1} is exactly the opposite of that in Figure \ref{fig:appendix_badcase1}. In this case, although both the graph index method and the flat index method retrieve the required session in the end, the noise from the retrieved chunks in the flat method interferes with generating the correct final result. In contrast, the session retrieved by the graph index method has less interference.

In Figure~\ref{fig:appendix_goodcase2}, we observe a clear advantage of the graph index when handling complex, multi-session data. In this scenario, the high retrieval recall facilitated by the graph's topological awareness successfully translates into accurate QA results. Conversely, the flat index fails to capture the long-range dependencies required for the query, leading to incomplete answer.

\begin{figure*}[t]
\centering
\begin{tcolorbox}[
    colback=white, 
    colframe=black!75, 
    fonttitle=\bfseries,
    title={Case Study: Graph Success case(ID: 7a87bd0c)},
    arc=0mm,
    boxrule=0.8pt,
    width=1.0\textwidth,
    left=2mm, right=2mm, top=2mm, bottom=2mm
]
    \small
    \textbf{User Query:} \textit{``How long have I been sticking to my daily tidying routine?''} \\
    \textbf{Answer:} 4 weeks.\\
    \textbf{Ground Truth Session:}\\
    1. "user":I need help organizing my garage this weekend. Can you give me some tips on how to sort through all the boxes and storage bins? Oh, and by the way, I've been feeling really proud of myself for sticking to my daily tidying routine - it's already been \textbf{3 weeks}! "assistant":...\\
    2."user":I'm planning to clean out the garage this weekend, but I need some advice on how to organize all the storage bins and boxes. Do you have any tips? By the way, I've been sticking to my daily tidying routine for\textbf{ 4 weeks now}, and it's amazing how much of a difference it's made in my apartment."assistant"...

    \vspace{2mm}
\begin{tabularx}{\textwidth}{X | X}
        \rowcolor{gray!10}
        \centering \textbf{Graph Index} & \centering \textbf{Flat Index} \tabularnewline
        \midrule
        \textbf{Retrieved Sessions:} & \textbf{Retrieved Sessions:} \\
        $\bullet$ \textbf{Session 1 (Target):} \textit{Date: 2023/06/11.} ...proud of myself for sticking to my daily tidying routine - it's already been \textbf{3 weeks}! & $\bullet$ \textbf{Session 1 (Target):} \textit{Date: 2023/06/11.} ...sticking to my daily tidying routine - it's already been \textbf{3 weeks}! \\
        $\bullet$ \textbf{Session 2-4:} [Irrelevant: Discussions on chicken stir-fry recipes, Thai green curry, and oven temperature tips.] & $\bullet$ \textbf{Session 2-4:} [Irrelevant: Detailed recipes for stir-fry and curry; kitchen design inspiration.] \\
        $\bullet$ \textbf{Session 5 (Target):} \textit{Date: 2023/09/30.} ...I've been sticking to my daily tidying routine for \textbf{4 weeks} now, and it's amazing... & $\bullet$ \textbf{Session 5 (Target):} \textit{Date: 2023/09/30.} ...sticking to my daily tidying routine for \textbf{4 weeks} now, and it's amazing how much... \\
        \addlinespace[1mm]
        \hdashline
        \addlinespace[1mm]
        {\color{green!40!black} \textbf{Model Response (Success):}} & {\color{red!70!black} \textbf{Model Response (Failure):}} \\
        \textit{``To answer the question... it is likely that the routine started around \textbf{4 weeks} before the current date, which is September 30, 2023.''} & \textit{``To answer the question...Based on the information extracted, I can see that the user has been sticking to their daily tidying routine for \textbf{at least 3 weeks, and possibly 4 weeks}, depending on the session...it is clear that the user has been sticking to their daily tidying routine for at least 3 weeks.''} \\
    \end{tabularx}
\end{tcolorbox}
\vspace{-8pt}
\caption{\textbf{Case where the Graph Index succeeds in knowledge update task while the Flat Index fails.} In this instance, both strategies successfully retrieve the necessary sessions. While the narrative coherence of the flat index was advantageous in Figure \ref{fig:appendix_badcase1}, this case demonstrates that such an advantage is not absolute.}
\label{fig:appendix_goodcase1}
\end{figure*}

\begin{figure*}[t]
\centering
\begin{tcolorbox}[
    colback=white, 
    colframe=black!75, 
    fonttitle=\bfseries,
    title={Case Study: Graph Success case (ID: 5025383b)},
    arc=0mm,
    boxrule=0.8pt,
    width=1.0\textwidth,
    left=2mm, right=2mm, top=2mm, bottom=2mm
]
    \small
    \textbf{User Query:} \textit{``What are the two hobbies that led me to join online communities?''} \\
    \textbf{Answer:} Photography and Cooking.\\
    \textbf{Ground Truth Session:} \\
    1. "user":I'm looking for some advice on camera lenses.... for street \textbf{photography}. By the way, I've been really enjoying editing my photos in Lightroom - the online communities I've joined have been super helpful in learning new techniques and getting feedback on my work....\\
    2. "user":I'm looking for some recipe inspiration...I've already joined a few online communities related to \textbf{cooking}, which led me to engage ...
    
    \vspace{2mm}
    \begin{tabularx}{\textwidth}{X | X}
        \rowcolor{gray!10}
        \centering \textbf{GraphIndex} & \centering \textbf{FlatIndex} \tabularnewline
        \midrule
        \textbf{Retrieved Sessions:} & \textbf{Retrieved Sessions:} \\
        $\bullet$ \textbf{Session 1 (Target):} \textit{``I'm looking for some advice on camera lenses....''} & $\bullet$ \textbf{Session 1:} \textit{``I'm looking for some book recommendations...''} \\
        $\bullet$ Session 2: I need help with creating a content calendar for my social media posts... & $\bullet$ Session 2: What are some effective ways to build social connections and overcome loneliness?... \\
        $\bullet$ Session 3: What is the current state of the real estate market in New York City... & $\bullet$ Session 3: How does Doug stay up-to-date with his hobbies or interests... \\
        $\bullet$ \textbf{Session 4 (Target):} \textit{``I'm looking for some recipe inspiration...''} & $\bullet$ \textbf{Session 4 (Target}): I'm looking for some recipe inspiration... \\
        $\bullet$ Session 5: I'm looking for some advice on finding rare collectibles online... & $\bullet$ Session 5: I'm looking for some advice on finding rare collectibles online... \\
        \addlinespace[1mm]
        \hdashline
        \addlinespace[1mm]
        {\color{green!40!black} \textbf{Model Response (Success):}} & {\color{red!70!black} \textbf{Model Response (Failure):}} \\
        \textit{``To answer the question...I can see that the user mentioned two hobbies that led them to join online communities: Photography (Session 1) Cooking (Session 4)...Therefore, the answer to the question is:...photography and cooking.''} & \textit{``To answer the question...From the chat history, I can see that the user mentioned the following hobbies: Reading books (Session 1, Session 4) Cooking (Session 4, Session 5)...Therefore, the two hobbies... are:1. **Reading books**2. **Cooking**''} \\  
    \end{tabularx}
\end{tcolorbox}
\vspace{-8pt}
\caption{\textbf{Case where the Graph Index succeeds in multi session task while the Flat Index fails.} In this instance, the query requires aggregating evidence across disjoint and non-contiguous sessions. The \textbf{Graph Index} accurately recalls both the photography-related (Session 1) and cooking-related (Session 4) contexts. Conversely, the \textbf{Flat Index} fails to retrieve the initial photography session, leading the model to rely on irrelevant retrieved noise and subsequently hallucinate ``reading books'' as a hobby. This comparison highlights the robustness of graph-based indexing in maintaining high recall for long-range, cross-session dependencies.}
\label{fig:appendix_goodcase2}
\end{figure*}

\subsection{Case Study: Entity Name vs. Entity Description as Keys}
To further investigate the impact of different indexing strategies on retrieval performance, we conduct a qualitative analysis using a temporal reasoning query. As illustrated in Figure \ref{fig:appendix_description_case}, the user asks about a specific past action: "I mentioned cooking something for my friend a couple of days ago. What was it?"

\textbf{Limitations of Entity Name as Key:} When using the entity name as the primary retrieval key, the system tends to retrieve broad, high-level concepts. For instance, entities like "RECIPE IDEAS" or "MEAL PREP IDEAS" are successfully identified, but their associated content remains too generic—focusing on the user's general goals rather than specific past events. Furthermore, temporal entities like "COUPLE OF DAYS" are retrieved as abstract durations without being grounded to the actual context of "baking." Consequently, the model fails to locate the specific "chocolate cake" event, leading to a response that only mentions general meal preferences.

\textbf{Advantages of Entity Description as Key:} In contrast, using the entity description as the key allows the retriever to leverage the semantic richness of summarized events. In this case, the system successfully retrieves the entity "LAST WEEKEND" because its description explicitly contains the specific detail: "A time period when the user made a chocolate cake for a friend's birthday party." By indexing the description, the semantic gap between the user’s query ("cooking something for a friend") and the stored memory is bridged more effectively.

\textbf{Conclusion:} This case study demonstrates that for complex long-term memory tasks—especially those involving temporal reasoning and specific event recall—entity descriptions serve as more informative keys than entity names. Descriptions capture the unique "who, what, and when" of an interaction, whereas names often fall back on redundant or overly-categorized labels that lack the granularity required for precise retrieval.

\begin{figure*}[t]
\centering
\begin{tcolorbox}[
    colback=white, 
    colframe=black!75, 
    fonttitle=\bfseries,
    title={Case Study: Precision in Temporal and Event Retrieval (ID: 9a707b82)},
    arc=0mm,
    boxrule=0.8pt,
    width=1.0\textwidth,
    left=2mm, right=2mm, top=2mm, bottom=2mm
]
    \small
    \textbf{User Query:} \textit{``I mentioned cooking something for my friend a couple of days ago. What was it?''} \\
    \textbf{Answer:} A chocolate cake.\\
    \textbf{Ground Truth Session:}\\
    \textit{Date: 2022/03/15.} ...I'm excited to try making croissants again... By the way, I just baked a \textbf{chocolate cake} for my \textbf{friend's birthday party last weekend} that turned out amazing. It was a new recipe I found online...\\
    \vspace{2mm}
\begin{tabularx}{\textwidth}{X | X}
        \rowcolor{gray!10}
        \centering \textbf{Entity Name as Key} & \centering \textbf{Entity Description as Key} \tabularnewline
        \midrule
        \textbf{Retrieved Entities \& Context:} & \textbf{Retrieved Entities \& Context:} \\
        $\bullet$ \textbf{Entity:} ``COUPLE OF DAYS'' (Duration) & $\bullet$ \textbf{Entity:} ``LAST WEEKEND'' (Time) \\
        \textit{Content:} A short period of time, approximately 2-3 days. & \textit{Content:} \textbf{A time period when the user made a chocolate cake for a friend's birthday party.} \\
        $\bullet$ \textbf{Entity:} ``RECIPE IDEAS'' (Goal) & $\bullet$ \textbf{Entity:} ``DINNER PARTY'' (Event) \\
        \textit{Content:} The user seeks innovative dessert and savory dish suggestions. & \textit{Content:} A social gathering hosted by the user to showcase their culinary skills. \\
        $\bullet$ \textbf{Entity:} ``MEAL PREP IDEAS'' (Behavior) & $\bullet$ \textbf{Entity:} ``ROASTED VEGETABLES'' (Object) \\
        \textit{Content:} User is seeking meal preparation ideas that can be reheated. & \textit{Content:} A meal prep item made by the user last weekend. \\
        \addlinespace[1mm]
        \hdashline
        \addlinespace[1mm]
        {\color{red!70!black} \textbf{Model Response (Failure):}} & {\color{green!40!black} \textbf{Model Response (Success):}} \\
        Based on your previous mentions, you were looking for \textbf{meal prep ideas} and \textbf{recipe ideas} for a dinner party, but there is no specific mention of what you cooked for a friend a few days ago. & You mentioned that \textbf{last weekend} you baked a \textbf{chocolate cake} for your friend's birthday party. You noted it was a new recipe using espresso powder. \\
    \end{tabularx}
\end{tcolorbox}
\vspace{-8pt}
\caption{\textbf{Comparison of retrieval performance between Entity Name and Entity Description as keys.} The description-based key successfully links the temporal entity ``Last Weekend'' to the specific event of ``baking a chocolate cake,'' whereas the name-based key retrieves generic categories.}
\label{fig:appendix_description_case}
\end{figure*}

\section{Prompts} \label{appendix_prompts}

Figure \ref{fig:prompt_entity_relation_extraction} shows the prompt we use for graph construction.
We use the official prompts from LongMemEval to extract memory (summary, factual statements and keywords).
To generate the answers and judge the results, we use the official prompts from LongMemEval or HaluMem in the corresponding experiments.
We use the default answering setting (CON + JSON) in LongMemEval experiments.

\newtcolorbox{promptbox}[2][]{%
    colback=gray!5!white,      
    colframe=gray!75!black,    
    fonttitle=\bfseries,       
    title={#2},                
    sharp corners,             
    boxrule=0.8pt,             
    left=6pt, right=6pt, top=6pt, bottom=6pt, 
    #1
}

\begin{figure*}[!t]
    \centering
    \begin{promptbox}{Prompt for Entity Relation Extraction}
        \small
        \textbf{Goal:}
        Given a multi-turn conversation consisting only of the user's messages (each turn separated by "\textbackslash n"), extract structured information that reflects the user's activities, possessions, goals, behaviors and reactions. Identify all relevant entities and their relationships to build a knowledge graph representing the user's context and life events.

        \vspace{0.5em}
        \textbf{Steps:}
        \begin{compactenum}
            \item Treat the entire conversation as one continuous narrative reflecting the user's life. Integrate information across all turns to infer complete and coherent entities and relationships.
            
            \item Identify all entities mentioned or implied by the user. For each entity, extract:
            \begin{compactitem}
                \item \texttt{entity\_name}: Name of the entity, capitalized.
                \item \texttt{entity\_type}: One of the following types: [User, Person, Object, Resource, Event, Goal/Intention, Time, Statistic, Duration, Place, Organization, Interest/Skill, Sentiment, Health, Behavior, Other]
                \item \texttt{entity\_description}: A comprehensive description summarizing how this entity relates to the user and any attributes mentioned (e.g., purpose, frequency, purchase time, emotional tone).
            \end{compactitem}
            Format each entity as:
            \begin{verbatim}
("entity"<|><entity_name><|><entity_type><|><entity_description>)            \end{verbatim}
            
            \item \textbf{Time Normalization and Extraction:}
            Whenever a specific or relative date is mentioned in the conversation, standardize it as a separate entity of type \texttt{"time"}.  
            Follow these rules:
            \begin{compactitem}
                \item Use the provided conversation time \texttt{\{dialogue\_time\}} as reference.
                \item If an explicit date is mentioned (e.g., "March 2nd"), convert it to \texttt{YYYY/MM/DD} format.
                \item If a relative time (e.g., "yesterday", "last week") appears, infer its absolute date relative to \texttt{\{dialogue\_time\}}.
                \item Do \textbf{not} create separate entities for recurring or habitual times (e.g., "every morning", "three times a week"); include such patterns only in related entity/relationship descriptions.
                \item Each time entity should describe \textbf{what happened at/before/after that time}.
            \end{compactitem}
            
            \item \textbf{Quantitative \& Frequency Extraction:}
            Explicitly extract any quantity, count, frequency, or duration mentioned in the conversation that describes the user's actions, achievements, or possessions.
            Include these as separate \texttt{"Statistic"} or \texttt{"Duration"} entities.
            Examples:
            \begin{verbatim}
("entity"<|>"Three Goals"<|>"Statistic"<|>"The user has scored 3 goals...")
("entity"<|>"Three Times A Week"<|>"Statistic"<|>"The user performs an activity...")
("entity"<|>"Five Weeks"<|>"Duration"<|>"The activity lasted for 5 weeks.")            \end{verbatim}
            
            \item From the identified entities, detect all pairs of (source\_entity, target\_entity) that have a meaningful or causal relationship in the context of the user's life. For each relationship, extract:
            \begin{compactitem}
                \item \texttt{source\_entity}: name of the source entity
                \item \texttt{target\_entity}: name of the target entity
                \item \texttt{relationship\_description}: a natural-language description explaining the relationship or connection between the source entity and the target entity.
                \item \texttt{relationship\_strength}: a numeric score (1–10) estimating how strong or explicit this connection is.
            \end{compactitem}
            Format each relationship as:
            \begin{verbatim}
("relationship"<|><source_entity><|><target_entity>
 <|><relationship_description><|><relationship_strength>)            \end{verbatim}
            
            \item Return output in English as a single list of all identified entities and relationships. Use \textbf{\#\#} as the list delimiter.
            
            \item When finished, output \texttt{<|COMPLETE|>}.
        \end{compactenum}
        \#\#\#\#\#\#\#\#\#\#\#\#\#\#\#\#\#\#\#\#\#\#\#\#\#\#\#\#\#\\
        \vspace{.5em}
        \textbf{Real Data:}\\
        \#\#\#\#\#\#\#\#\#\#\#\#\#\#\#\#\#\#\#\#\#\#\\
        \begin{verbatim}
Conversation time: {dialogue_time}
Text: {input_text}
        \end{verbatim}
        \#\#\#\#\#\#\#\#\#\#\#\#\#\#\#\#\#\#\#\#\#\#\\
        \vspace{0.5em}
        \textbf{Output:}
    \end{promptbox}
    \caption{The specific prompt used for entity and relation extraction}
    \label{fig:prompt_entity_relation_extraction}
\end{figure*}

\end{document}